\documentclass[journal]{IEEEtran}

\usepackage{cite}
\usepackage{amsmath}
\usepackage{footnote}
\usepackage{bm}
\usepackage{amssymb}
\usepackage{amsfonts}
\usepackage{graphicx}
\usepackage{float}
\usepackage{subfig}
\usepackage{mathtools}
\usepackage{booktabs}
\usepackage{dutchcal}
\usepackage{xcolor}
\usepackage{tabularx}
\usepackage{array}
\usepackage{algorithm}
\usepackage{algpseudocode}
\usepackage{url}
\usepackage{multirow} 
\usepackage{verbatim}

\begin{document}

\title{Divide et Impera: Decoding Impedance Strategies for Robotic Peg-in-Hole Assembly}

\author{
Johannes~Lachner,\thanks{J. Lachner, F. Tessari, M.C. Nah, and N. Hogan are with the Department of Mechanical Engineering, Massachusetts Institute of Technology, Cambridge, MA, USA A.M. West Jr. is with the Department of Mechanical Engineering, Johns Hopkins University, Baltimore, MD, USA; e-mail: jlachner@mit.edu, ftessari@mit.edu, awest36@jh.edu, mosesnah@mit.edu, neville@mit.edu}
Federico Tessari,
A. Michael West Jr., 
Moses C. Nah, 
Neville Hogan\thanks{J. Lachner and N. Hogan are also with the Department of Brain and Cognitive Sciences, Massachusetts Institute of Technology, Cambridge, MA, USA}
}

\maketitle

\begin{abstract}
This paper investigates robotic peg-in-hole assembly using the Elementary Dynamic Actions (EDA) framework, which models contact-rich tasks through a combination of submovements, oscillations, and mechanical impedance. Rather than focusing on a single optimal parameter set, we analyze the distribution and structure of multiple successful impedance solutions, revealing patterns that guide impedance selection in contact-rich robotic manipulation.
Experiments with a real robot and four different peg types demonstrate the presence of task-specific and generalized assembly strategies, identified through K-means Clustering. Principal Component Analysis (PCA) is used to represent these findings, highlighting patterns in successful impedance selections. Additionally, a neural-network-based success predictor accurately estimates feasible impedance parameters, reducing the need for extensive trial-and-error tuning. By providing publicly available code, CAD files, and a trained model, this work enhances the accessibility of impedance control and offers a structured approach to programming robotic assembly tasks, particularly for less-experienced users.
\end{abstract}


\section{Introduction}\label{sec:Intro}

A current trend in robotic control is the development of contact-rich manipulation techniques \cite{nagabandi2020deep,andrychowicz2020learning,chen2022system,seo2023contact,pang2023global,xie2023neural}. Despite the existence of many effective methods for robot motion planning, managing the physical interaction of robots with various objects and surfaces remains a significant research challenge. However, mastering contact-rich scenarios is crucial for robots to achieve human-like interaction capabilities, such as those required in industrial assembly lines.

Lately, robotics has seen promising developments in contact-rich manipulation, such as model-based methods \cite{posa2014direct,manchester2020variational,pang2023global} and learning-based approaches \cite{oikawa2021reinforcement,chen2022system,Chi_2023_Diffusion,xie2023neural}. Model-based methods leverage mathematical representations of contact dynamics, ensuring that the robots can predict and react to contact forces accurately. Learning-based methods leverage probabilistic models to manage uncertainties in contact interactions. In particular, approaches such as reinforcement learning and neural networks allow robots to develop optimal contact strategies through trial and error, fostering adaptable manipulation skills.

Contact-model-based optimization approaches integrate detailed contact models into the control framework, considering factors such as friction cones, contact modes, and complementarity constraints. These models are crucial for tasks requiring precise interaction with objects, such as manipulation in cluttered environments. Techniques like Mixed Integer Quadratic Programming, which treat contact modes as integer variables \cite{marcucci2017approximate,hogan2020reactive}, and Linear Complementarity Problems \cite{anitescu1997formulating, posa2014direct, aydinoglu2022real}, are commonly used. However, these methods face significant challenges, including numerical complexity, sensitivity to coordinate frame choice and object geometry, and the need for accurately estimated parameters like friction coefficients.

Sampling-based optimization methods have also gained traction, particularly in scenarios where gradient information is not readily available. Most of these sampling-based methods are Zeroth-order methods, which do not rely on gradients and sample multiple actions and update control parameters based on outcomes without needing derivative information \cite{theodorou2010generalized}. Variations such as STOMP \cite{kalakrishnan2011stomp}, PI² \cite{buchli2011learning,stulp2012reinforcement}, and MPPI \cite{williams2017model,abraham2020model,van2023learning} have become popular. First-order methods, involving randomized smoothing for gradient-based optimization, are also employed \cite{suh2022bundled,suh2022differentiable,pang2023global}. 
The approaches stochastically smooth the underlying landscape of the objective function to enhance robust planning and control \cite{duchi2012randomized,curtis2012sequential,burke2020gradient}.

The methods discussed above often rely on predefined cost functions. These functions aim to identify a single “optimal” set of control parameters. They frequently converge to trivial solutions, such as uniformly increasing all parameters, resulting in behavior resembling position controllers~\cite{buchli2011learning, Manschitz_2020}. While multiple feasible solutions are common, their distribution and interdependence have note been not investigated in these approaches.

Other  approaches in contact-rich robotic control are Dynamic Movement Primitives (DMP) and Elementary Dynamic Actions (EDA). Both approaches are inspired by human motor control research \cite{schaal1999imitation, Ijspeert_2013, hogan2012dynamic,nah2024MotorPrimitives} and try to leverage insights into how humans achieve complex, coordinated movements \cite{kim2011impedance,park2013intuitive}. It is hypothesized that keys to human performance are building blocks of coordination, as originally proposed by Sherrington \cite{sherrington1906integrative}. These building blocks are conventionally called “motor primitives” and are basic motor actions that can generate a rich repertoire of behaviors. Bernstein further expanded on this concept by introducing the idea of “repetition without repetition” \cite{bernstein_1967}. He observed that humans can accomplish tasks in numerous ways, adapting to subtle environmental changes and learning strategies for movement. This adaptability is a key ability of the human motor system. Latash argued that the motor system does not operate with explicit costs but rather utilizes a subspace of multiple solutions, enabling flexible and effective movement strategies \cite{Latash_2012}. Wolpert et al. emphasized the importance of motor learning, highlighting that feed-forward control is essential for natural movements, as feedback mechanisms can be too slow\cite{wolpert1998multiple, Wolpert_2001}. 

The main motivation of the work presented here is to demonstrate that contact-rich robotic tasks generally do not have a single optimal solution; rather, multiple parameterizations can successfully achieve the desired performance. Drawing inspiration from principles well-established in human motor control, we propose that there exists a set of viable solutions capable of adequately performing a given task. To explore this, we adopt a robot control framework based on Elementary Dynamic Actions (EDA) \cite{hogan2012dynamic, Lachner_2023_EDA, nah2024MotorPrimitives, Nah_2023_EDA_kin}, aiming to achieve adaptability comparable to human motor performance. Although EDA has demonstrated notable success in contact-rich manipulation tasks \cite{fasse1996control, Lachner_2023_EDA}, effectively parameterizing its dynamic primitives—particularly stiffness and damping—remains a key challenge.

\begin{figure}[h]
    \centering
\includegraphics[width=\columnwidth, trim={0.0cm 0.0cm 0.0cm 0.0cm}, clip]{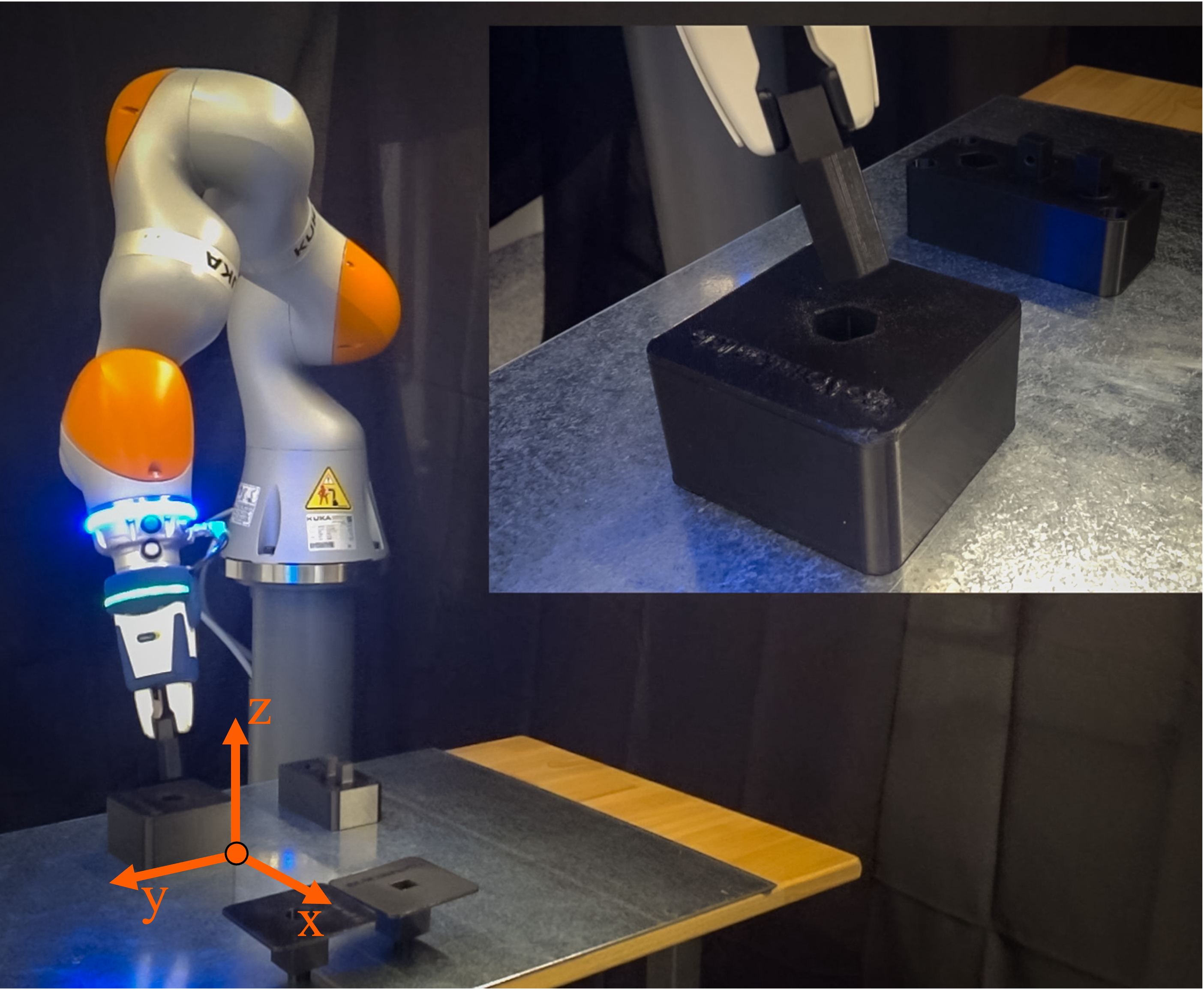}
  \caption{Exemplary peg-in-hole assembly using a KUKA LBR robot. All control parameters were defined with respect to the spatially fixed coordinate frame highlighted in orange.}
  \label{fig:AssemblyApp}
  \vspace{-0.5cm}
\end{figure}

This paper examines peg-in-hole assembly, a foundational task in robotics (see Fig.~\ref{fig:AssemblyApp}). Peg-in-hole assembly serves as a representative application for numerous industrial robotic tasks, showcasing essential aspects of contact-rich manipulation. Examples include assembling engine components in the automotive industry, connecting circuit board plugs in electronics manufacturing, and future human-centric applications like furniture assembly.

We present a comprehensive investigation into the influence of impedance parameters, such as stiffness and damping, on task success in robotic peg-in-hole assembly. Our contributions are:
\begin{enumerate}
    \item Through Principal Component Analysis (PCA) and K-means Clustering, we reveal two key insights: (a) task-specific strategies tailored to individual peg types and (b) generalized strategies applicable across various peg shapes.
    \item We develop a neural network model to predict successful impedance parameters, providing a practical tool to support robotic programmers, particularly to aid those with limited experience.
\end{enumerate}

\bigskip
All hardware and software tools used in this work are publicly available, to enable researchers to test our findings, extend our dataset, or even use the software for real peg-in-hole assembly applications in industry.

\section{Methods}\label{sec:Methods}
This section outlines the theoretical framework, experimental protocol, and computational techniques employed to investigate the role of impedance parameters in robotic peg-in-hole assembly. Throughout the remainder of the paper, we consider a system with $n$ degrees of freedom (DOF).

\subsection{Impedance Control}\label{subsec:ImpedanceControl}
Impedance control \cite{hogan1984impedance, hogan1985impedance} models the robot controller as a Norton equivalent network \cite{hogan2014general}, defining the relationship between generalized force output $\mathbf{F} \in se^{\star}(3)$ and motion input $\dot{\mathbf{x}} \in se(3)$ at the interaction port $(\mathbf{F}, \dot{\mathbf{x}})$ (Fig.~\ref{fig:EDA}B). The controller generates interaction forces $\mathbf{F}$ by coupling the difference between a virtual trajectory $\mathbf{x}_0 \in SE(3)$ and the current robot pose $\mathbf{x} \in SE(3)$ with mechanical impedance $\mathbf{Z}: se(3) \rightarrow se^{\star}(3)$. Modulating $\mathbf{Z}$ allows control over the dynamics of physical interaction. For example, tactile exploration and manipulation of delicate objects benefit from low stiffness, while tasks like drilling demand high stiffness \cite{hogan2018impedance}. Under the assumption that the environment is an admittance, mechanical impedance can be linearly superimposed even though each mechanical impedance is a nonlinear operator \cite{hogan1985impedance,hogan2017physical}. For $i$-number of impedances: $\mathbf{Z} = \sum \mathbf{Z}_i$. More information about impedance control can be found in \cite{hogan1984impedance, hogan1985impedance, lachner2022geometric}.

\subsection{Elementary Dynamic Actions (EDA)}\label{subsec:EDA}
\begin{figure*}[htp]
    \centering
  \includegraphics[width=0.95\textwidth, trim={0.0cm 0.0cm 0.0cm 0.0cm}, clip]{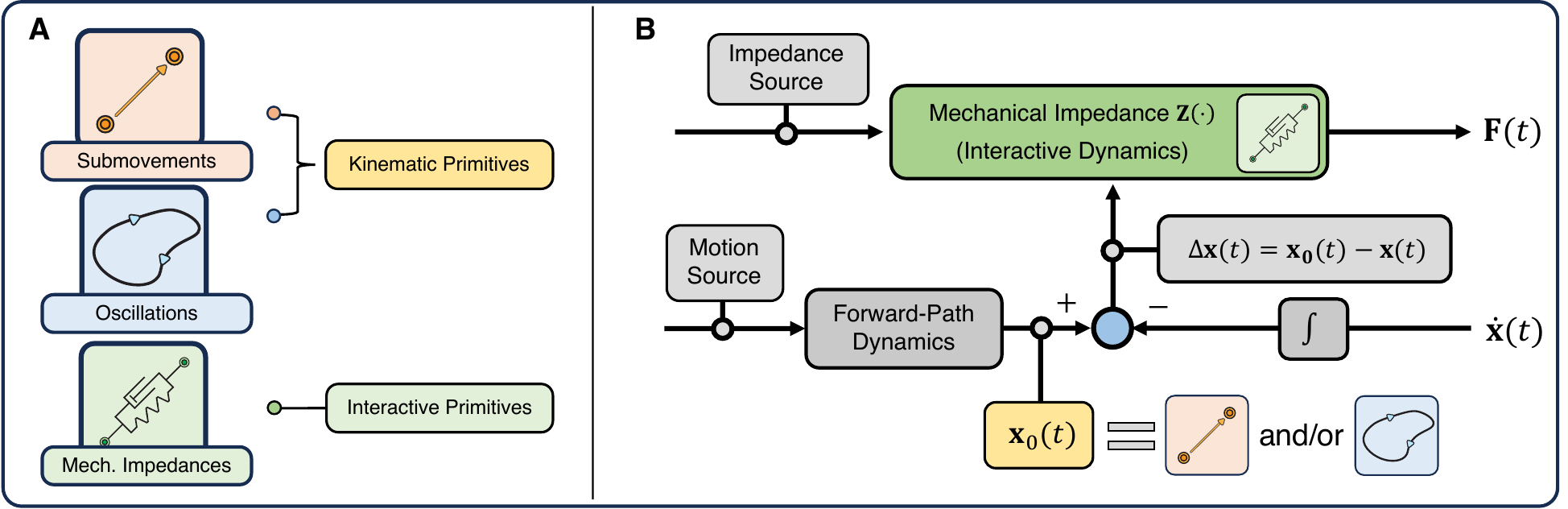}
  \caption{(A) The three primitives of Elementary Dynamic Actions (EDA). Submovements and oscillations are kinematic primitives used to generate discrete and periodic movements respectively \cite{Nah_2023_EDA_kin}, and mechanical impedance is a dynamic primitive used to handle physical interaction \cite{hogan2017physical, hogan2018impedance,hogan2022contact, lachner_energy_2021, lachner_shaping_2022}. (B) The three primitives are combined using a Norton equivalent network model \cite{hogan2014general}.}
  \label{fig:EDA}
  \vspace{-0.4cm}
\end{figure*}

EDA relates impedance control to principles of human motor control by defining the virtual trajectory $\mathbf{x}_0$ through two key motor primitives: discrete movements and oscillatory motions \cite{Nah_2023_EDA_kin, nah2024MotorPrimitives}. Additionally, it incorporates mechanical impedance as a distinct dynamic primitive to manage physical interaction \cite{hogan2017physical, hogan2012dynamic, hogan2013dynamic}. These three primitives are integrated within a Norton equivalent network model (Fig.~\ref{fig:EDA}B).  

\subsubsection{Submovement}\label{subsec:submovement}
A submovement $\mathbf{x}_0$ is a smooth trajectory in which its time derivative is a unimodal function, i.e., has a single peak value:
\begin{equation}
    \dot{\mathbf{x}}_0(t) = \mathbf{v} \ \hat{\sigma}(t)
\end{equation}
In this equation, $\hat{\sigma}:\mathbb{R}_{\ge 0}\rightarrow [0,1]$ denotes a smooth unimodal basis function with peak value 1; 
$\mathbf{v}\in\mathbb{R}^{n}$ is the velocity amplitude of the submovement. 
Submovements model discrete reaching motions, and therefore $\hat{\sigma}(t)$ has a finite support, i.e., there exists $T>0$ such that $\hat{\sigma}(t)=0$ for $t\ge T$. 
\cite{hogan2007rhythmic}

\subsubsection{Oscillation}\label{subsec:oscillation}
An oscillation $\mathbf{x}_0$ is a smooth non-zero trajectory which is a periodic function:
\begin{equation}
    \forall t >0: ~~ \exists T>0: ~~\mathbf{x}_0(t) = \mathbf{x}_0(t+T)
\end{equation}
Compared to submovements, oscillations model rhythmic and repetitive motions.

\subsubsection{Mechanical Impedance}\label{subsec:mechanical_impedance}
Mechanical impedance serves as a dynamic primitive in the EDA framework to regulate physical interactions between the robot and its environment. As discussed in Sec.~\ref{subsec:ImpedanceControl}, mechanical impedance $\mathbf{Z}$ defines the relationship between the displacement $\Delta \mathbf{x}$ (difference between virtual and actual trajectories) and the resulting force $\mathbf{F}$. An overview of the literature on mechanical impedance in human motor control can be found in \cite{Hermus2023}.

\subsection{Assembly strategy for peg-in-hole tasks}\label{subsec:AssemblyStrategy}
For our assembly application, we used a KUKA LBR robot and a Cartesian Impedance Controller \cite{lachner2022geometric}. The robot application code is provided on Github\footnote{\url{https://github.com/mosesnah-shared/ImpedanceClassification/tree/main/Sunrise}}.

We programmed an application to assemble four different peg-in-hole types: square, triangular, \sloppy hexagonal, and cylindrical pegs. Between the workpieces and the respective insertion holes was a nominal clearance of 0.14 mm (square peg), 0.18 mm (hexagonal peg), and 0.20 mm (circular and triangular pegs). All pieces were printed with a PRUSA i3 MK3 3D-printer using PLA filament. All components except the pegs were printed with $20 \%$ infill with gyroid pattern. Pegs were printed with higher infill ($> 70 \% $) to guarantee structural strength. The CAD data of all pieces can be found on Dropbox\footnote{\url{https://www.dropbox.com/scl/fo/ljrkha4bh17v4scy8uy9l/AL4zqjOkfwap6kvq8QBylWQ?rlkey=qqyxf8euipigzhlokfo5x2nr6&dl=0}}.

Algorithm 1 shows the structure of the assembly application. For our application, we used the torque sensors of the robot to detect external contact (Algorithm 1, line 3-5 and 14-20). 
\begin{algorithm}
    \caption{ Primitives for peg-in-hole assembly: submovements (orange), oscillations (blue), and mechanical impedances (green). The coordinates related to the algorithm can be seen in Fig.~\ref{fig:AssemblyApp}. The impedance parameters in line 11-13 are exemplary and were varied throughout the experiments. The frame-based discrete movements and oscillations remained fixed throughout the experiments.}\label{alg:assemblyStrategy}
    \begin{algorithmic}[1]
        \State Incline workpiece and align planarly at center of hole
        \State Create spatial \texttt{Force Condition}: $F_{\text{TCP}} < 15$ N
        \While{\texttt{Force Condition}}
                \State \colorbox{orange}{Move} \texttt{linRel} along $-x$ and $y$ 
        \EndWhile
        \State \colorbox{orange}{Rotate} from current TCP-pose to final TCP-pose, with\\
            \hspace{0.5cm} \colorbox{green}{Siffness}:\\
            \hspace{0.5cm} \texttt{CartDOF.TRANSL} $= 400$ [N/m]\\
            \hspace{0.5cm}\texttt{//Be compliant about A\\
            \hspace{0.2cm} //for peg to ``slip'' into the hole}\\
            \hspace{0.5cm} \texttt{CartDOF.A} $= 5$ [Nm/rad]\\
            \hspace{0.5cm} \texttt{CartDOF.B,C} $= 100$ [Nm/rad]\\
            \hspace{0.5cm} \colorbox{green}{Damping} \texttt{CartDOF.ALL} $= 0.7$
        \While{ \texttt{Force Condition} }
            \State \colorbox{orange}{Move} \texttt{linRel} along $-z$, with \State \colorbox{green}{Siffness} (line \texttt{8-12}), \colorbox{green}{Damping} (line \texttt{13})\\
            \hspace{0.5cm} \texttt{//Handle riples of 3D-print}
            \State Overlay \colorbox{cyan}{Oscillation} about \texttt{CartDOF.A}, with
            \State Amplitude $A= 0.6$ [N], Frequency $f = 4.5$ [Hz]
        \EndWhile 
    \end{algorithmic} 
\end{algorithm}

The overall assembly strategy was inspired by observations of humans performing assembly tasks, often without visual feedback. Humans frequently rely on tactile feedback and subtle movements, such as wiggling the workpiece, to align and insert components accurately. 

For the peg-in-hole assembly, we chose a fixed strategy for the zero-force trajectory, $\mathbf{x}_0$, and investigated the influence of different choices of impedance parameters, $\mathbf{Z}$. For the kinematic primitives (submovements and oscillations), the frame-based programming procedure of KUKA Sunrise was used. This design choice allowed us to focus our analysis solely on the solution space of impedance parameters, excluding variations in the kinematic properties of the zero-force trajectory. Notably, prior work has explored the inverse approach, i.e., varying the kinematic primitives with constant dynamic primitives \cite{nah2020dynamic,nah2023learning}. For the impedance parameters, the diagonal terms of the translational and rotational stiffness matrices and the translational and rotational damping ratios were varied throughout the experiments.

With the assembly strategy (Algorithm~\ref{alg:assemblyStrategy}), the peg automatically ``slipped'' into the hole, even though the workpiece was not perfectly aligned during the assembly \cite{lachner2023elementary}. This is especially beneficial for end-of-line assembly where the workpiece is not perfectly positioned in front of the robot. Expensive fixation mechanisms can  be obviated.

Due to imperfections of the 3D-print, the rippled surfaces of the hole and the workpiece could impede the insertion of the peg. We therefore overlaid small oscillations on the discrete motion along $-z$ during the assembly (Algorithm~\ref{alg:assemblyStrategy}, line $18-19$). While certainly other assembly strategies would be successful for this application (e.g., combining position control and perception), for us the key was the combination of tactile exploration with kinematic primitives and impedance. This demonstrated the benefits of the modularity of EDA. A video of the assembly strategy can be found on YouTube\footnote{\url{https://youtu.be/T2pFu6IXk04}}.

As can be seen in algorithm~\ref{alg:assemblyStrategy}, many parameters must be selected. The exact number of parameters depends on the assembly strategy of the programmer. In this paper, we emphasized  the impedance parameter choice (green). Thus, for our experiments we fixed the parameters of the kinematic primitives (submovements and oscillations in algorithm~\ref{alg:assemblyStrategy}). With this, a set of eight impedance values remained to be selected.  The scalar stiffnesses along and about the coordinates $\{ x, y, z \}$ are denoted as $\{ k_x, k_y, k_z \}$ and $\{ k_C, k_B, k_A \}$, respectively. Note that the notation is defined in this way by the KUKA Sunrise programming interface. For example, $k_C$ represents the rotation about the $x$-axis . The common damping ratio values for translational and rotational motion are denoted $\{\zeta_t, \zeta_r \}$. More information about the controller parameterization can be found in the KUKA Sunrise manual\footnote{\url{https://my.kuka.com/s/product/kuka-sunriseos-117/}}. 

\subsection{Experimental protocol}

\begin{figure}[h]
    \centering
  \includegraphics[width=\columnwidth, trim={0.0cm 0.0cm 0.0cm 0.0cm}, clip]{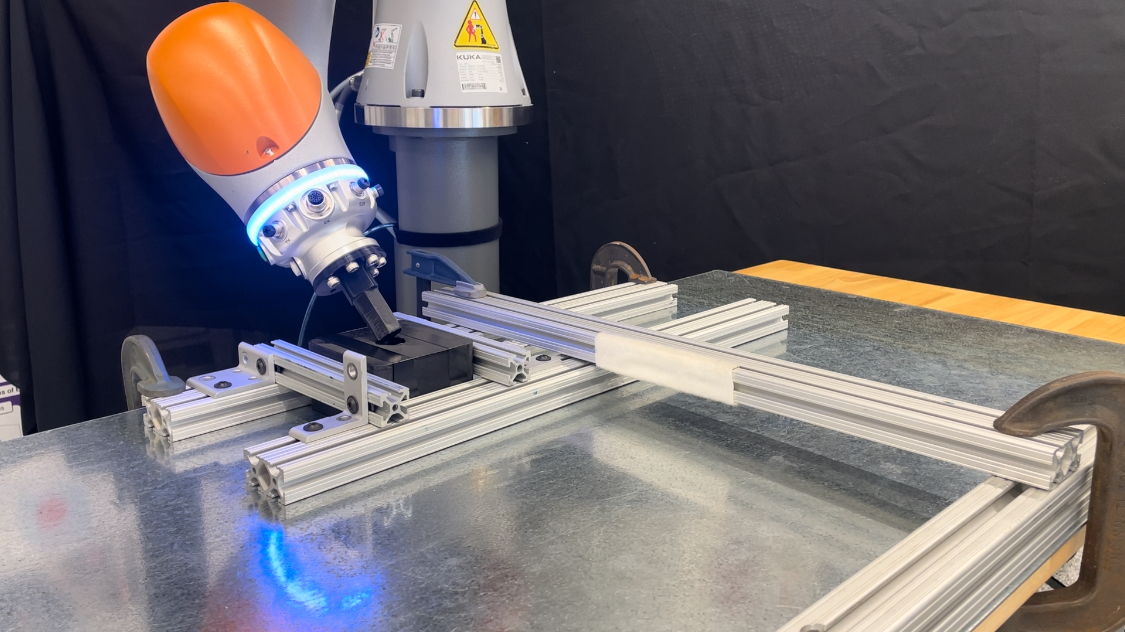}
  \caption{Experimental setup used during the trials. The workpiece with the hole was secured to prevent any movement throughout the four 40-hour long experimental trials.}
  \label{fig:Experiments}
\end{figure}

The control parameters that can realistically be rendered by the robot were  determined to be in a range of $50 \leq k_t \leq 1000$ N/m for  translational stiffnesses, $5 \leq k_r \leq 200$ Nm/rad for the rotational stiffnesses, and $0.1 \leq \zeta \leq 0.9$ for damping ratio. These ranges were found empirically. 

Finding feasible impedance values is not trivial. Our intention was to test combinations of impedance values for peg-in-hole assembly. Due to the curse of dimensionality \cite{Bellman_1957_DynamicProgramming}, it was not practical to test all possible combinations of impedance values. Hence, we assigned three different categories for each of the parameter ranges: small, medium, and large. The exact values of the three categories can be found in Appendix~\ref{ap:Categories}. 

For each impedance parameter, we randomly selected a value from one of the three categories totalling $3^8 = 6561$ impedance combinations tested for each peg (Fig.~\ref{fig:Experiments}). As each iteration required an average of 22 seconds, the total experimental time for each peg took about 40 hours. For four pegs, this gave an overall experimental time of about 160 hours.

\subsection{Computational methods}\label{subsec:Comp_methods}
After conducting experiments, we investigated key aspects of successful peg-in-hole assembly:
\begin{itemize}
    \item The success rates associated with different impedance parameter combinations.
    \item The variability of assembly strategies tailored to individual peg types and their low-dimensional representation.
    \item The identification of common strategies, applicable across various peg shapes, represented in a low-dimensional space.
\end{itemize}
The methods used to derive these findings are detailed in the following subsections.
\vspace{2mm}  

\subsubsection{Linear regression to fit the success rate}
A total of $N_{tot} = 3^8 = 6561$  impedance parameter combinations was tested for each peg shape. The success rate, $s_r$, was computed as the percentage of successful trials, $N_s$, over the number of total tested parameters, $N_{tot}$.

\begin{equation}
    s_r = 100\cdot \frac{N_s}{N_{tot}}
\end{equation}

To understand how different impedance parameter choices affected the outcome of the peg-in-hole task, we used linear regression to fit the success rate (dependent variable) as a function of the impedance parameters (independent variables) for each peg. A linear model was fit to each success rate distribution for each impedance parameter and peg, and the goodness-of-fit was assessed using the coefficient of determination ($R^2$), which measures the variance in the dependent variable that is predictable from the independent variable. The resulting linear fit was statistically tested against the degenerate null slope, testing the null hypothesis of no relationship between the success rate and the impedance parameters. For example, a significant positive slope in the relationship between success rate and the stiffness $k_z$ implied that higher vertical stiffness tended to improve task success.

\vspace{2mm}  
\subsubsection{PCA to identify a low dimensional representation}
To reduce the complexity of high-dimensional data and reveal (potential) hidden patterns in data, we used PCA to project the data onto a lower dimensional subspace, often referred to as Principal Components (PCs). The data were pre-processed using the algorithm proposed by West et al. \cite{west2023}, i.e., the input dataset was centered with respect to its mean thus removing an otherwise implicit bias in the first principal component.
Moreover, a significant reduction in dimension, from eight to three or fewer, would enable us to visualize the 8-dimensional solution space in the lower-dimensional PC space. 

We applied PCA to the impedance solutions obtained from our peg-insertion task to test if  a non-arbitrary lower-dimensional subspace could account for our data. We projected the data into a subspace defined by the PCs that captured the most variance. The  main PCs provide insight into the principal directions in the data. To extract PCs from the impedance solutions, we utilized the \texttt{pca} function from MATLAB 2022b.

To verify that the observed solution space reduction was not random, we compared the variance explained by the first three dominant PCs to that of an equally-sized random subset of trials. More detailed information about this statistical test can be found in Appendix~\ref{ap:PCA}.

Finally, we projected both successful and unsuccessful trials into the identified PC subspace to visually assess their distribution. If unsuccessful trials consistently fell outside the identified subspace, this would further support the non-arbitrary nature of the impedance solution subspace.

\vspace{2mm}  
\subsubsection{K-means Clustering to reveal different assembly strategies}
We utilized K-means Clustering on successful trials within our eight-dimensional impedance solution space to investigate clusters that represent distinct common impedance solution strategies. We employed MATLAB 2022b’s K-means Clustering algorithm with an online phase update and five replicates \cite{lloyd1982least, arthur2006k}. The online updates, wherein each sample was individually reassigned to a new cluster if it reduced the sum of point-to-centroid distances, ensured convergence to a local minimum. The use of multiple replicates added robustness to the clustering process. More details about the clustering process can be found in Appendix~\ref{ap:SilhouetteCoeff}.

Since PCA-based dimensionality reduction removes variance that may contain important structure within the impedance space, clustering was performed on the original eight-dimensional space to fully capture all impedance parameter relationships. However, for better visualization, we projected the resulting clusters onto a 2D or 3D PCA-reduced space, allowing for an intuitive representation while maintaining the integrity of the clustering process.

To assess the distinctiveness of the identified clusters, we conducted a one-way MANOVA. Details of this statistical validation can be found in Appendix~\ref{ap:MANOVA}.

\vspace{2mm}  
\subsubsection{Neural network to reveal common assembly strategies}
To investigate common assembly strategies applied to all pegs, we used a neural network for binary classification. Our goal was to find the best-fit model that predicted success for our experimental data with $N = 4 \cdot 3^8$ data points and labels $y = \{ 0,1 \}$, i.e., true (1) and false (2). The ``best-fit” was determined by the minimum of the Binary Cross Entropy:
\begin{equation}
    BCE = - \sum^{N}_{i=1} \{ y_i\log(\hat{y}_i) + (1-y_i)\log(1-\hat{y}_i) \}.
\end{equation}
We used a binary cost, as  only success or failure is important for  assembly, not the minimization of a continuous error function such as the Euclidean error.

For given input impedance parameters $\bm{x} = \{k_x, k_y, k_z, k_C, k_B, k_A, \zeta_t, \zeta_r \}$, we wanted to find the function $f(\bm{x})$ that represents the boundary of our data, dividing it into success or failure. 

The network was trained for all pegs, i.e., the eight impedance values and their success flags were merged into one dataset. Once the network was trained, we used the network to produce successful data and analyze whether all pegs shared a common assembly strategy, represented by a lower-dimensional PC space (Sec.~\ref{subsec:CommonAssembly}). More details about the neural network structure can be found in Appendix~\ref{ap:NeuralNetwork}.

\section{Results}
This section presents the findings of our investigation into the role of impedance parameters in robotic peg-in-hole assembly. We highlight task-specific and generalized strategies derived through PCA and K-means Clustering. We introduce a neural network model designed to predict and visualize successful impedance parameters.

\subsection{Success rate and distribution}\label{subsec:SuccessRate}

\subsubsection{Success Rate}
\begin{table}[h]
\centering
\caption{Percentage success rate and total number of successes for the different tested pegs.}
\label{tab:success_rate}
\begin{tabular}{@{}ccc@{}}
\toprule
Peg      & Success Rate {[}\%{]} & Total Success \\ \midrule
Triangle & 1.54                  & 101/6561 \\
Square   & 9.85                  & 646/6561\\
Hexagon  & 2.76                  & 181/6561\\
Cylinder & 2.24                  & 147/6561\\ \bottomrule
\end{tabular}
\end{table}

Tab.~\ref{tab:success_rate} presents the success rates for each peg, along with the absolute number of successful insertions. The square peg achieved the highest success rate ($9.9\%$), while the other three pegs exhibited significantly lower success rates, ranging from $1.5 \%$ to $2.8 \%$. We hypothesize that the square peg’s higher success rate is due to its use in the initial programming phase, where kinematic primitives—such as translations, rotations, and oscillations—were optimized and remained unchanged for the other peg shapes.

The overall low success rates highlight the difficulty of tuning impedance parameters for contact-rich tasks, emphasizing the challenge faced by robot programmers in identifying feasible parameter sets. Despite these challenges, our results show that among the $6,561$ tested impedance parameters, a considerable number of successful solutions were found, ranging from $101$ for the triangular peg to $646$ for the square peg, reinforcing the importance of structured impedance selection strategies in robotic assembly.

Interestingly, despite the complexity of impedance tuning, our findings demonstrate that multiple successful impedance solutions exist for each peg shape, even when a fixed assembly strategy is applied uniformly across all pegs. This suggests that rather than a single optimal impedance value, there is a broader range of viable solutions, complementing optimization methods that typically focus on identifying a single best set of impedance parameters \cite{buchli2011learning, abu2020variable}.

\subsubsection{Success Rate Distribution}

\begin{figure*}[t]
\includegraphics[width=\textwidth]{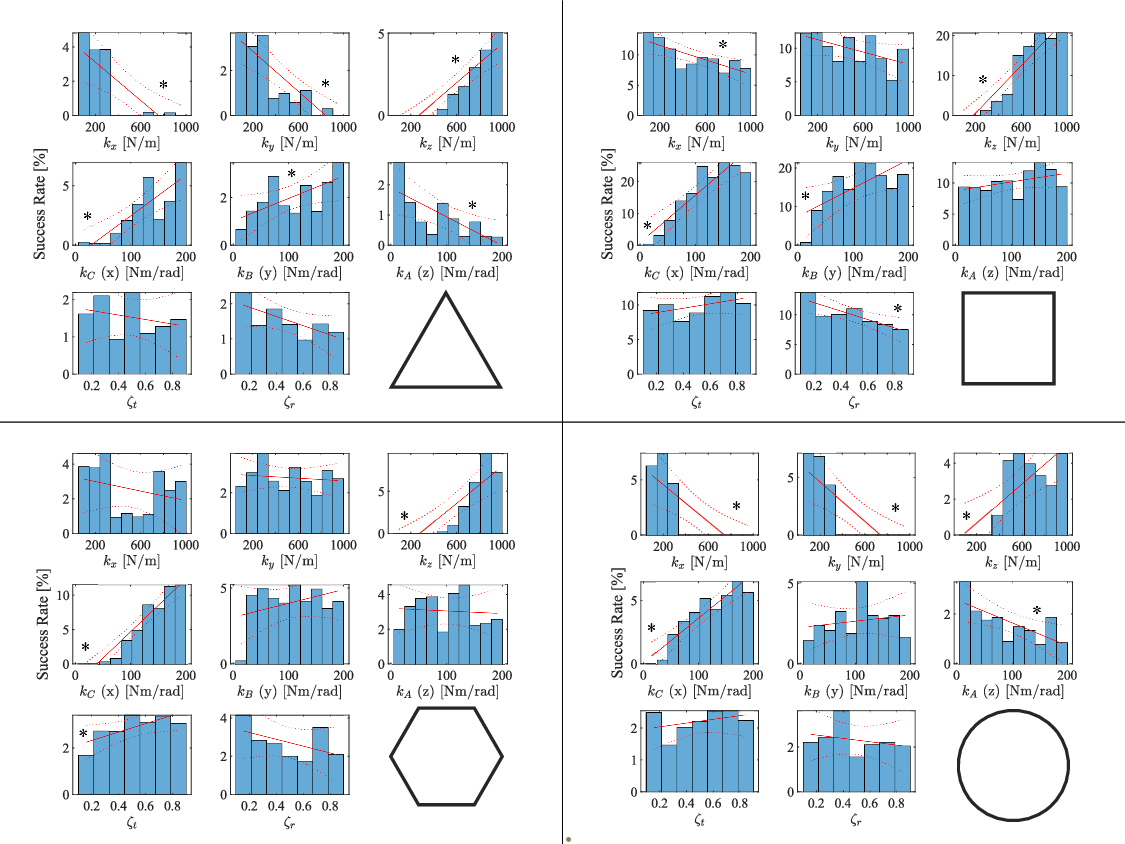}
\centering
\caption{Distribution of success rates (blue bars) for each impedance parameter for the four different pegs (triangle, square, hexagon, and cylinder). For each histogram, the red solid lines represent the best linear fit, while the red dashed lines show the confidence intervals. An asterisk (*) was added for those linear fits that presented a significant slope.}
\label{fig:relative_succ_rate}
\end{figure*}

For each of the pegs, the distribution of success rates for each impedance parameter ($k_x, k_y, k_z, k_C, k_B, k_A,\zeta_t,\zeta_r$) was computed by dividing each impedance parameter into a number of equally spaced bins and computing the success rate for each bin. 

Fig.~\ref{fig:relative_succ_rate} shows the observed distributions. Interestingly, different pegs shared similar success rate distributions. The success rate grew with increase in vertical stiffness $k_z$ in all pegs, and the linear models presented good coefficients of determination ($R^2_{triangle}=0.86,R^2_{square}=0.91,R^2_{hexagon}=0.77,R^2_{cylinder}=0.65$) with significant slopes ($p<0.05$) for all pegs. A similar behavior was observed for the success rate and the $k_C$
rotational stiffness across all pegs. The linear models presented high coefficients of determination \sloppy($R^2_{triangle}=0.73,R^2_{square}=0.85,R^2_{hexagon}=0.94,R^2_{cylinder}=0.88$) and significant slopes ($p<0.05$) for all pegs.

Another interesting finding was that the success rate appeared to be independent of the damping ratio parameters ($\zeta_t,\zeta_r$). In fact, linear fits of the success rate to these two parameters presented low coefficients of determination and non-significant slopes ($p > 0.05$) for all but two cases ($\zeta_{r_{square}},\zeta_{t_{hexagon}}$). 

The success rate for the other four impedance parameters ($k_x,k_y,k_A,k_B$) was less consistent  across pegs. Noticeably, the success rate distributions over $k_x,k_y$ in the triangle peg and the cylinder peg were similar, with a concentration of success for low values of these two impedance parameters. However, the same was not true for the square and hexagon pegs. 

Appendix~\ref{ap:LinearRegression} presents the complete details of the linear model results and fitting performance on the other four impedance parameters ($k_x,k_y,k_A,k_B$).

\subsection{Low dimensional representation of solutions}

To test whether there existed a non-arbitrary lower dimension subspace to represent the solutions, we projected the impedance solutions into a lower dimensional subspace using PCA. We found that 3 PCs could reasonably describe the data, a substantial reduction from the 8 dimensional solution space. Specifically, 3 PCs accounted for 83.8\%, 96.6\%, 95.9\%, 88.7\%, and 95.7\% of the successful trials of the cylinder, hex, square, triangle, and all peg shapes, respectively. 

To show that the reduction of the solution space was not a random artifact, we compared the slope of the variance-accounted-for by the first three principal components to the slope of the variance-accounted-for by a \textit{random subset} of $n$ trials from the data, where $n$ is the number solutions reported in Tab.~\ref{tab:success_rate}. 
The amount of variance-accounted-for by each PC in the solution space is shown Appendix~\ref{ap:PCA} (Fig.~\ref{fig:VAF_slope}).

We conducted a pair-wise t-test between the variance-accounted-for slopes of the successful trials and the random trials. We found that the slopes of the successful trials were significantly steeper; that is, the slopes of the successful trials were significantly more negative ($p < 0.001$). This confirms that the PCs extracted from the successful trials were not random artifacts; rather, they represent an impedance solution subspace.

\begin{figure*}[htp]
\includegraphics[width = \textwidth]{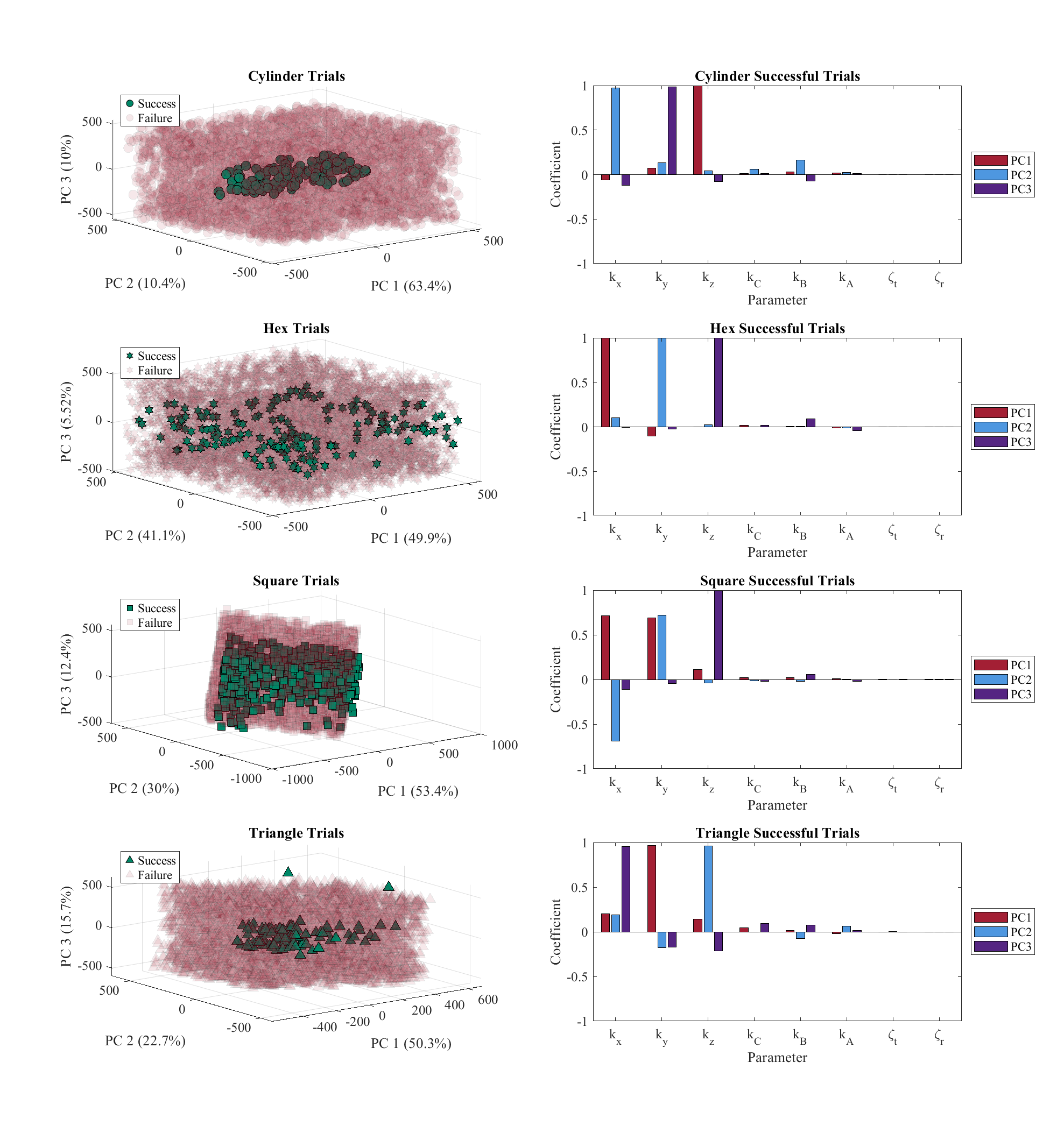}
\centering
\caption{\textbf{Left Column:} The successful (green) and unsuccessful (red) trials of the cylinder, hex, square, and triangle pegs are projected into the solution space obtained using PCA. Note that often the failures fall outside the PCA-reduced solution space. 2D projections of these graphs can be found in ~\ref{fig:SuccessvsFailure_combined_projections}. \textbf{Right Column: }The coefficients of each of the three Principal Components are shown.}
\label{fig:SuccessvsFailure_combined}
\end{figure*}

Because the data patterns were effectively captured by three PCs, we were able to visualize them in a 3-dimensional space without losing significant information. By projecting the failure trials into this 3-dimensional subspace, we observed that the failures frequently fell outside the identified solution subspace, particularly in the cylinder and triangle trials (Fig.~\ref{fig:SuccessvsFailure_combined}). This visualization underscores the existence of a meaningful, non-arbitrary lower-dimensional solution space to represent the impedance solutions. Moreover, analyzing this solution space highlighted the multitude of successful impedance parameter configurations within EDA, illustrating the flexibility and adaptability of this approach. This will be further elaborated in Sec.~\ref{subsec:CommonAssembly}.

\subsection{Task-specific assembly strategies for individual pegs}\label{subsection:DifferentStrategies}

\begin{figure*}[htp]
\includegraphics[width = \textwidth]{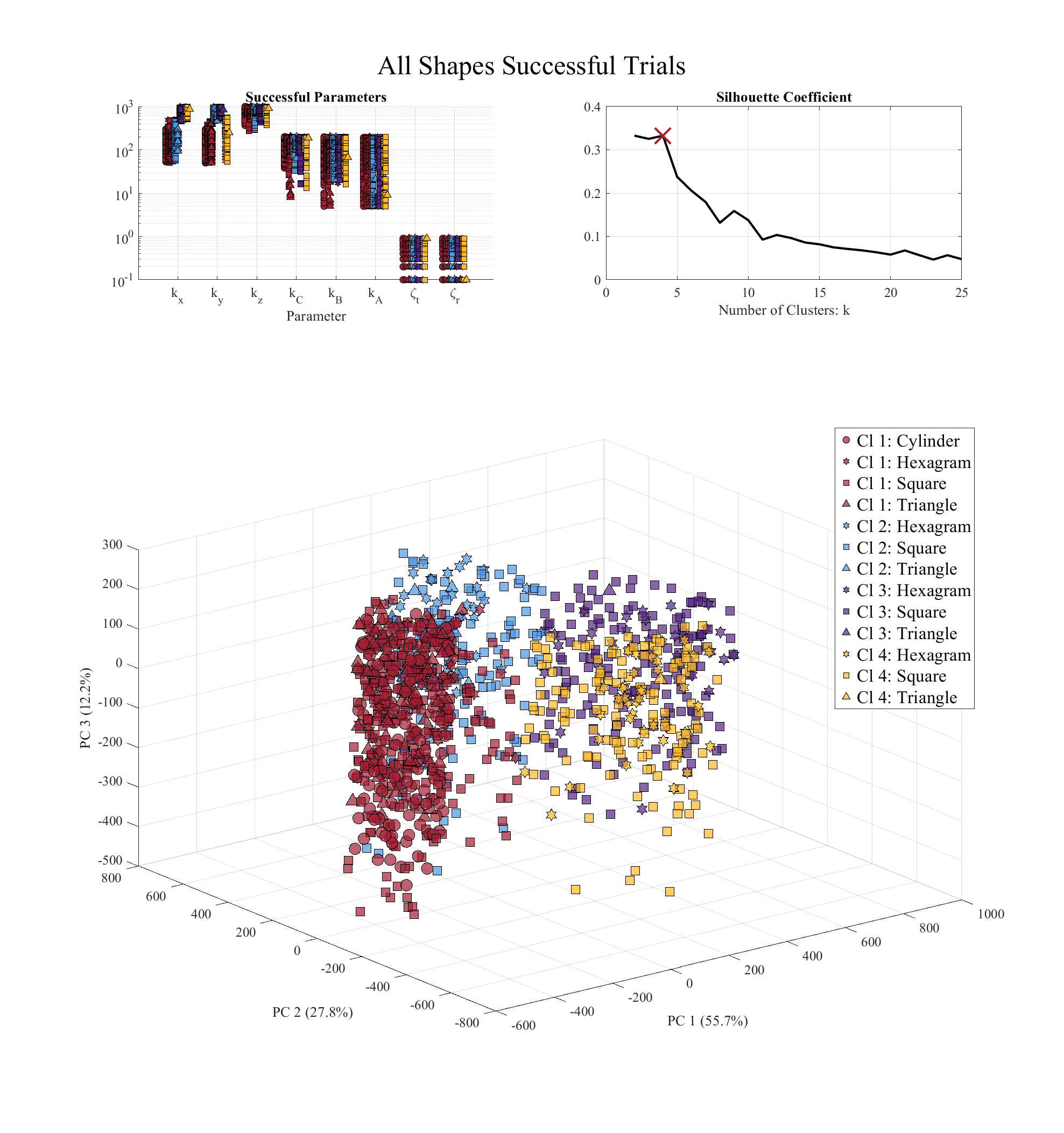}
\centering
\caption{\textbf{Top Left:} The successful parameters of all  pegs. \textbf{Top right:} The silhouette coefficient found from K-means Clustering as a function of the number of clusters. An optimal number of clusters is denoted by a red X. \textbf{Bottom: }The successful trials of all  pegs were projected into their solution space obtained using PCA. Each color denotes a different cluster identified by the k-means clustering algorithm. In the legend, ``Cl" is an abbreviation for cluster. Different marker shapes denote different pegs.}
\label{fig:ALL_kmeans}
\end{figure*}

We used K-means Clustering to identify clusters of different successful assembly strategies in our eight-dimensional impedance solution space. Specifically, we used the silhouette coefficient to identify the ``optimal'' number of clusters. In Fig.~\ref{fig:ALL_kmeans} we see that the optimal number of clusters was four, with a silhouette coefficient of 0.33. Further, in Fig.~\ref{fig:ALL_kmeans} we see that often multiple clusters contained different pegs. Only Cluster 1 (red) contained all four pegs; it was the only cluster to include cylinder solutions. Together, these results suggest that similar strategies enabled different tasks (i.e., different pegs).

To verify the distinctiveness of the identified clusters, we conducted a one-way Multivariate Analysis of Variance (MANOVA), using the cluster group as the independent variable with four levels. The results revealed significant differences among the clusters on the combined dependent variables, indicating that the multivariate means of the clusters spanned at least three dimensions (Appendix~\ref{ap:MANOVA_combined}). These findings validate the clusters identified by K-means Clustering, confirming distinct multivariate mean vectors for each cluster. This demonstrates the existence of task-specific assembly strategies for individual pegs. We identified families of solutions, revealing 2 clusters for the cylinder peg, 4 for the hex peg, 3 for the square peg, and 2 for the triangle peg (Fig. \ref{fig:SilhouetteCoefficent}). More information about their silhouette coefficients can be found in appendix~\ref{ap:SilhouetteCoeff_pegTypes}.

\begin{figure*}[htp]
\includegraphics[width = \textwidth]{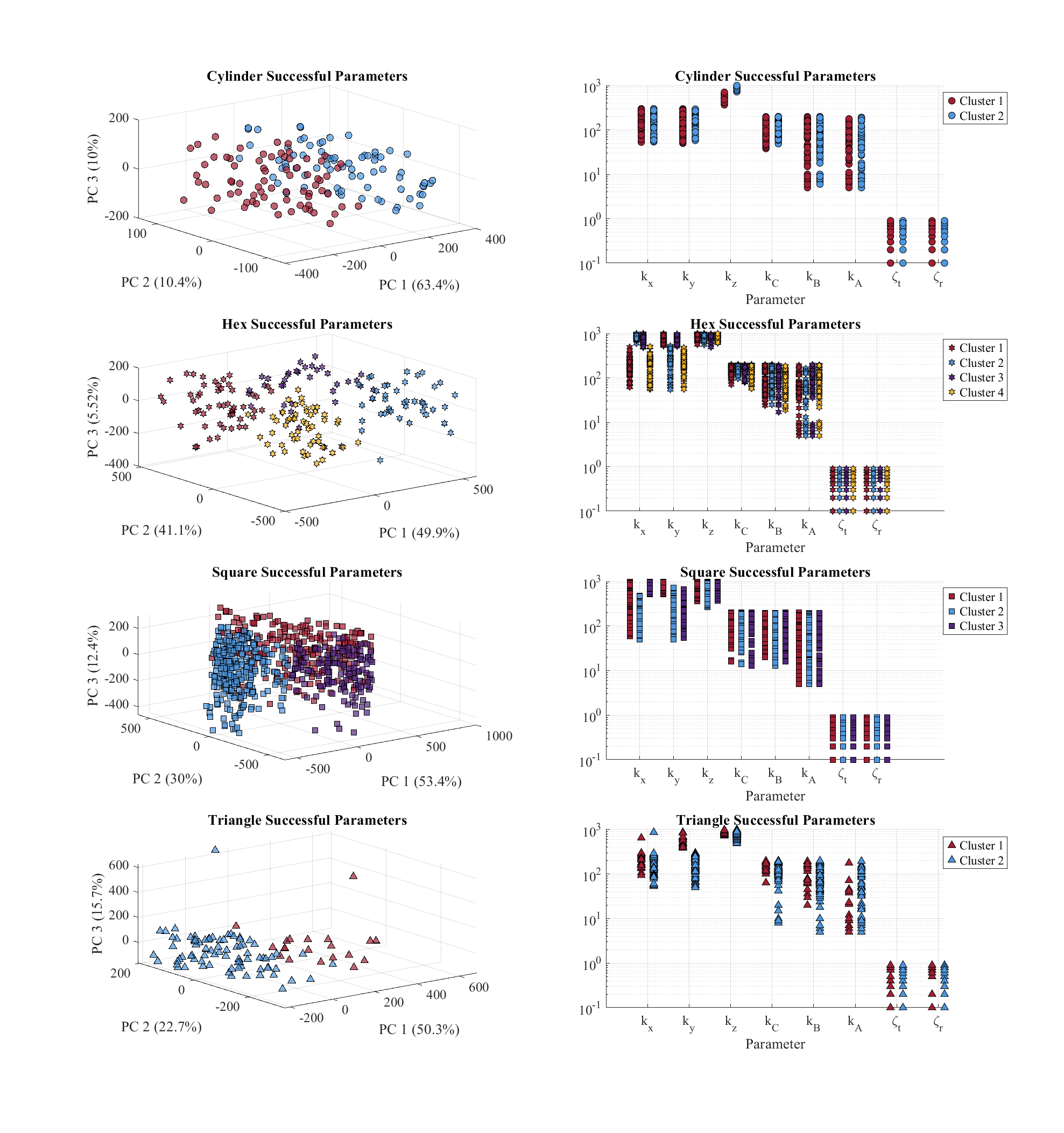}
\centering
\caption{\textbf{Left Column:} The successful trials of each peg are projected into their solution space obtained using PCA (see Fig.~\ref{fig:SuccessvsFailure_combined}). Each color denotes a different cluster identified by the k-means clustering algorithm. \textbf{Right Column:} The successful parameters of each peg.}
\label{fig:COMBINED_kmeans}
\end{figure*}

The right column of Fig.~\ref{fig:COMBINED_kmeans} shows the different solution strategies that were identified. For example, in the successful trials of the cylinder peg, Cluster 1 contained lower values of $k_z$ while Cluster 2 contained higher values of $k_z$. The difference between the 4 clusters in the successful hex trials depended on the combination of $k_x$ and $k_y$: Cluster 1 contained lower values of $k_x$ and higher values of $k_y$, Cluster 2 contained higher values of $k_x$ and lower values of $k_y$, Cluster 3 contained higher values of both $k_x$ and $k_y$, and Cluster 4 contained lower values of both $k_x$ and $k_y$. The three clusters identified in the  successful square trials contained combinations of values in $k_x$, $k_y$, and $k_z$. Finally, the two clusters identified in the  successful triangle trials differed in their range of values in $k_y$. All in all, this indicates that through the constrained assembly strategy multiple combinations of $k_x$ and $k_y$ were feasible. This showcases that families of solutions exist which encode different assembly strategies.

To verify the distinctiveness of the identified clusters, we conducted a one-way Multivariate Analysis of Variance (MANOVA). The results confirmed significant differences among clusters for all peg shapes, indicating that the multivariate means of the clusters occupied distinct subspaces. For example, the cylinder and triangle clusters lay in one-dimensional spaces, the square clusters spanned two dimensions, and the hex clusters required three dimensions (appendix~\ref{ap:MANOVA_pegTypes}). These findings validate the distinctiveness of the solution clusters and support the conclusion that multiple families of solutions encode different assembly strategies.

\subsection{Generalized assembly strategies across all pegs}\label{subsec:CommonAssembly}

As we have shown in our results, a range of successful parameters exist for each peg (Sec.~\ref{subsec:SuccessRate}). These parameters might be grouped into different assembly strategies for each individual peg (Sec.~\ref{subsection:DifferentStrategies}). To examine whether a common assembly strategy existed across all pegs, we created a binary classification neural network to sample over the whole space of successful parameters (Sec.~\ref{subsec:Comp_methods}).

The neural network was created using PyTorch. We empirically determined $2^{13}$ training epochs to avoid over-fitting. The dataset was split into 70$\%$ training data and 30$\%$ test data. No cross-validation was used, as the dataset was large enough ($i = 4 \cdot 3^{8}$ for all four pegs) to inherently provide a good representation of the overall data space. A model accuracy of 95.92$\%$ was calculated based on the validation dataset.

\begin{figure*}[htp]
\includegraphics[width = \textwidth]{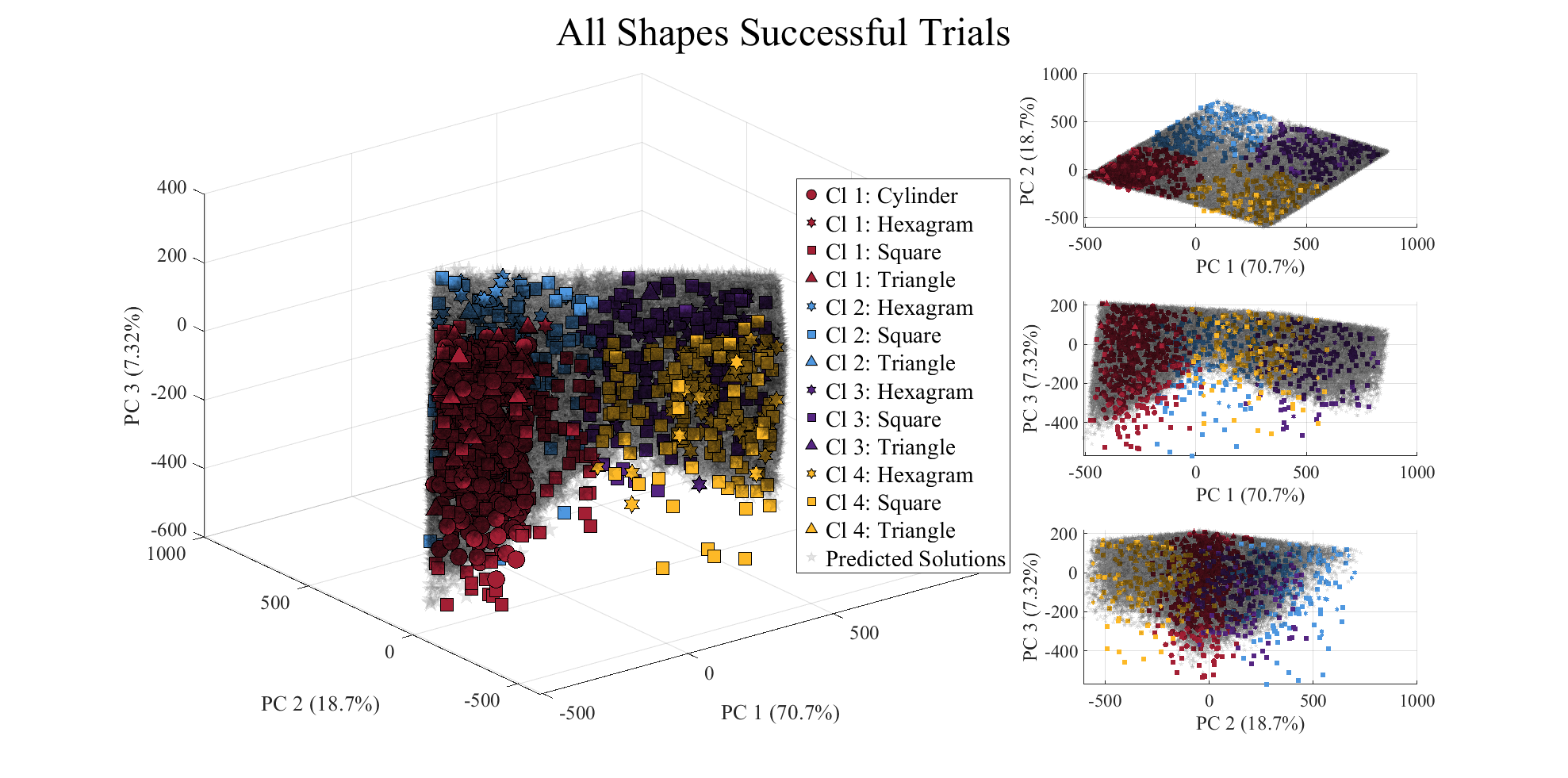}
\centering
\caption{\textbf{Left:} The successful parameters of all pegs (from Fig. \ref{fig:ALL_kmeans}) as well as the predicted solutions projected into a common PC space. \textbf{Right: }Various 2D views of the 3D image.}
\label{fig:predicted_success_data_kmeans}
\end{figure*}

We used the model above to produce 100,000 sets of successful impedance parameters; these data will be referred to as the predicted solutions. A PCA was conducted on a combined dataset of the 100,000 predicted solutions and the previously found successful trials. The projection into the first three PCs is shown in Fig.~\ref{fig:predicted_success_data_kmeans} (note that the clusters here are the same as those in Fig.~\ref{fig:ALL_kmeans}). Three dominant PCs were present, accounting for 96.7\% of the variance, while subsequent PCs accounted for less than 3\% each. 

Some interesting findings can be observed. First, the predicted solution space for all pegs (shown in grey) is in fact not a uniform cube, suggesting that there is a subspace---or a family of solutions---common to all pegs. Note that areas in the solution cube outside of this subspace suggest impedance solutions that are not common to all pegs. For example, Cluster 1 (red) contains a solution strategy encompassing all four pegs. The predicted solution space here contains a higher density of predicted solutions in this area (grey) as opposed to other areas (e.g., Cluster 4); this is best seen in the top right of Fig.~\ref{fig:predicted_success_data_kmeans}. This intuition is consistent with our results.

\begin{figure*}[htp]
\includegraphics[width = 0.7\textwidth]{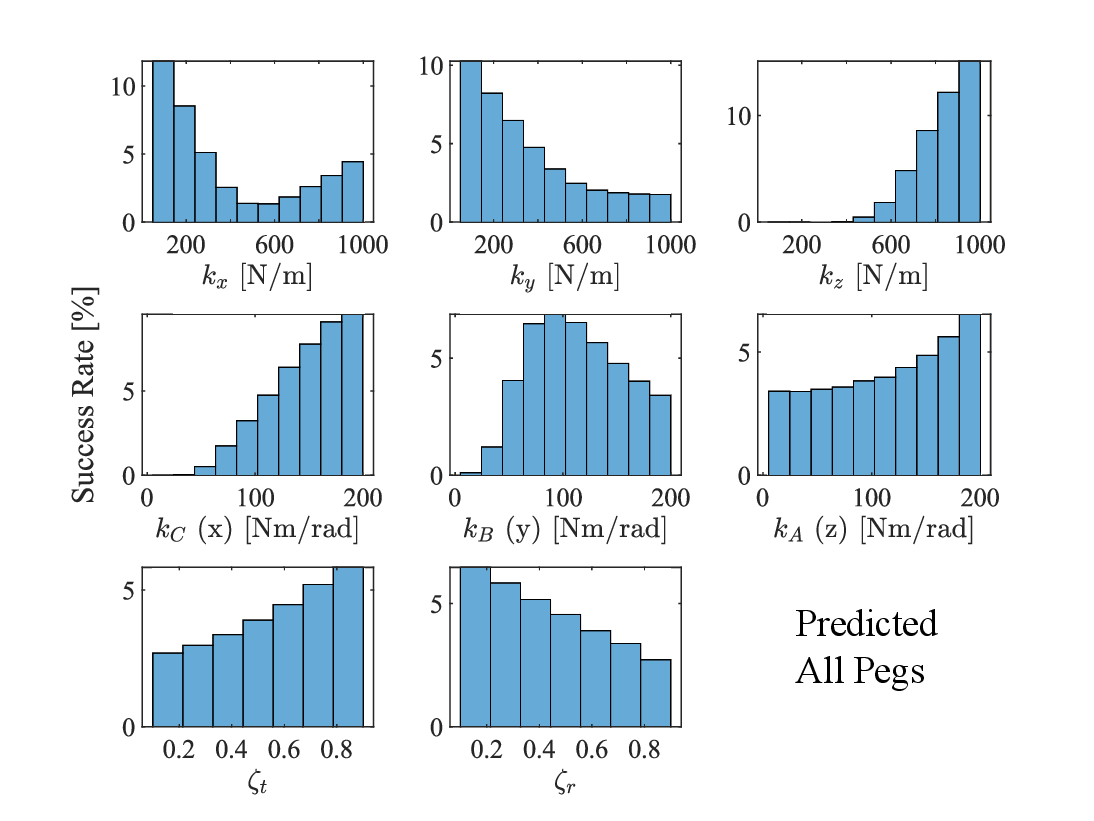}
\centering
\caption{Distribution of success rates for parameters predicted to be successful by the neural network. The distributions were generated by sampling impedance parameters uniformly between the minimum and maximum values specified in Sec.~\ref{sec:Methods} until 100,000 successes were achieved, resulting in a success rate of $4.32\%$.}
\label{fig:predicted_success_data_histograms}
\end{figure*}

Fig.~\ref{fig:predicted_success_data_histograms} presents the predicted success rates for all peg types, based on the same 100,000 data points. The predicted success rates for all individual pegs can be found in appendix~\ref{ap:LinReg_Neural}. 

Although it is challenging to interpret all solutions due to the neural network predicting outcomes across all pegs, some interesting trends can be observed. The success rate was highest for low linear stiffness values $k_x$ and $k_y$, suggesting that the peg more easily slips into the hole during assembly with these lower parameters, compensating for imperfect alignment. A similar trend to that observed in Fig.~\ref{fig:relative_succ_rate} is noted: success increases with higher vertical stiffness $k_z$ and rotational stiffness $k_C$. Once the peg is aligned with the hole, a certain force must be exerted, hence higher values of $k_z$ are more successful. Additionally, the original programmed orientation about the x-axis seemed to be similar across all pegs, which may explain the higher success rates for high rotational stiffness $k_C$. Interestingly, $k_A$ appears to be successful across all parameter ranges. The rotational stiffness $k_B$ shows the highest success rates at intermediate values, which might indicate that the final orientation of the peg and hole was not perfectly programmed and varied slightly across pegs. This could explain the optimal balance of maintaining orientation without being overly stiff about the y-axis. The damping parameters $\zeta_t$ and $\zeta_r$ were successful across all ranges, with a growing success rate for increasing values of $\zeta_t$ and a declining success rate for decreasing values of $\zeta_r$. Still, we conclude that for our assembly application, the effect of the selected damping ratio was minor, most likely due to the low speed of the task.

\subsection{Neural network-based impedance success predictor}\label{subsec:Application}

Even for a given assembly strategy (submovements and oscillations in algorithm 1),  determining  a  set  of  feasible  impedance  parameters $\{k_x, k_y, k_z, k_C, k_B, k_A, \zeta_t, \zeta_r \}$ is not trivial \cite{hogan1985impedance}. To address this challenge, we trained a neural-network-based success predictor (Sec.~\ref{subsec:CommonAssembly}) that can assist users—especially those working with peg geometries similar to ours—by providing immediate feedback on the feasibility of selected impedance parameters. 

To make this tool widely accessible, we provide an open-source Python implementation on GitHub\footnote{\url{https://github.com/mosesnah-shared/ImpedanceClassification/blob/main/ImpedanceClassification/success_failure_predictor.py}}. Users can input a set of impedance parameters, and the neural network will predict the probability of success, eliminating the need for time-consuming real-world testing. This is particularly valuable for inexperienced robot programmers, as it allows them to iteratively refine their parameter choices without direct physical experimentation.

To evaluate the neural network’s effectiveness, we randomly selected 30 predicted solutions and tested these parameter sets for each peg on the real robot. Remarkably, all parameter sets were successful, except for one case each for the square and triangle pegs. In total, these solutions resulted in 118 successful insertions out of 120 trials. A video of the experiments on the robot can be found on Dropbox\footnote{\url{https://www.dropbox.com/scl/fo/gav5l7kpzqs69dzbi06xj/AMGmfsqsbLDaThPr7xVikWo?rlkey=pc443f61levpdeme9exejhsa1&dl=0}}.

These results demonstrate that, despite the overall low success rate across the full parameter search space, the trained neural network significantly improved the success rate by guiding the selection of feasible impedance parameters. This reduces the need for exhaustive trial-and-error tuning, making impedance selection more systematic, efficient, and accessible for users.

\section{Discussion}\label{sec:Discussion}
Our study investigated key factors for successful peg-in-hole assembly, emphasizing the role of impedance control in contact-rich robotic manipulation. We demonstrated that a range of successful parameters exists, underscoring the flexibility of solutions rather than relying solely on optimization. This complements traditional approaches by suggesting multiple viable solutions rather than a single optimal one.

While conventional robotic control strategies often prioritize higher stiffness values to improve precision and stability, our findings suggest a different perspective. Lower stiffness values, particularly for parameters such as $k_x$ (translation along $x$), $k_y$ (translation along $y$), $k_A$ (rotation about $z$), $\zeta_t$ (translational damping ratio), and $\zeta_r$ (rotational damping ratio), proved effective. Physically, lower translational impedance facilitates easier insertion of the peg into the hole, suggesting that higher stiffness may not always be necessary or beneficial for optimal performance in contact-rich tasks.

Consistent with previous studies~\cite{Manschitz_2020, martin2019variable}, our research indicates that increasing stiffness along unconstrained directions, such as $k_z$, can indeed improve success rates. However, for other parameters, moderate or even lower stiffness values were equally effective, reinforcing the idea that parameter selection should be driven by task-specific characteristics rather than a single, universally optimal approach.

Interestingly, our results suggest that the damping ratio's influence on assembly success may be less critical than previously assumed. Thus, adopting default damping ratio values might be practical in many scenarios, simplifying parameter tuning without significantly compromising performance. However, more dynamic tasks with fewer physical constraints could necessitate more careful tuning of damping parameters.

\section{Conclusion}\label{sec:Conclusion}
Our study explored key aspects of successful peg-in-hole assembly, highlighting the advantages of impedance control through Elementary Dynamic Actions (EDA) for contact-rich robotic manipulation.

We demonstrated that EDA supports multiple successful impedance parameter sets, indicating flexibility and adaptability in selecting effective control strategies. Principal Component Analysis (PCA) revealed a structured, lower-dimensional representation of these successful solutions, simplifying the complexity of the impedance parameter space.

Notably, our findings highlight that certain impedance settings generalize across different peg types, revealing common underlying strategies. By categorizing impedance parameters into intuitive groups—low, medium, and high—we facilitate efficient programming and rapid adaptation to variations in assembly tasks. This approach aligns with Bernstein's \cite{bernstein_1967} concept of “repetition without repetition,” mirroring the adaptability of human motor control.

All CAD data and software related to this study are publicly available, providing researchers with valuable resources to validate, extend, or apply these findings to real-world robotic assembly applications.

\section{Limitations and future work}\label{sec:Limitations}
Our study also indicates that the effectiveness of EDA not only depends on the impedance parameter choice but also depends on the chosen strategy for kinematic primitives. While EDA’s modularity offers many benefits, the specific combination of submovements and oscillations used in our experiments played a crucial role in achieving successful outcomes. This dependency suggests that further refinement and customization of kinematic strategies may be necessary to optimize EDA’s performance across different tasks. Here, teaching by demonstration might be a promising way to facilitate the programming of kinematic primitives \cite{nah2024MotorPrimitives, Nah_2023_EDA_kin}.

Our study involved several parameter choices that could influence the outcomes of our investigation. For instance, during the Principal Component Analysis (PCA), we reduced the dimensionality of the solution space from eight to three principal components, capturing a certain percentage of the variance. This choice, while effective for our purposes, may not generalize across different contexts or datasets.

In our analysis, we employed K-means Clustering in conjunction with silhouette coefficients to ascertain the optimal number of clusters. However, it is important to acknowledge that the silhouette coefficient is fraught with several limitations. These limitations include its sensitivity to the shape and density of clusters, as it assumes clusters to be spherical and evenly sized, which may not always be the case in real-world data. In our case, the resulting silhouette coefficients were relatively low, ranging from 0.22 to 0.41, suggesting a possible absence of a distinct separation between clusters. This highlights one of the inherent challenges in clustering analysis: the difficulty of defining a universally accepted threshold for silhouette coefficients that reliably demarcates well-defined clusters. Determining the appropriate silhouette coefficient for effective cluster delineation remains an active and evolving area of research, which is beyond the scope of this paper.

In the neural network model for binary classification, we used a structure with two hidden layers, each consisting of 64 neurons, and trained the network over $2^{13}$ epochs. The Adam optimizer with a fixed learning rate of $1 \cdot 10^{-4}$ was used to update the weights. Although these are common choices in neural network design, other configurations and hyper-parameters could also have been effective in producing a feasible neural network.

An additional limitation of our study is the lack of coupling between degrees of freedom. Traditional solutions for peg-in-hole insertion, such as remote-center compliance (RCC), inherently couple the stiffness of different degrees of freedom to facilitate alignment and insertion tasks \cite{Whitney_1982, Hirai1990remote}. Our simplification to independent impedance parameters was made to keep the system accessible for novice programmers, aiming to maintain simplicity in implementation. However, it is important to acknowledge that this decoupling may overlook some of the benefits provided by more sophisticated compliance mechanisms. Future work should investigate the integration of coupled impedance parameters to enhance the robustness and accuracy of EDA in more complex assembly tasks.

Finally, while our results are promising for peg-in-hole assembly, further investigations are needed to evaluate how our results may be generalizable to other manipulation tasks. Different tasks may present unique challenges that require further investigation to determine the applicability and adaptability of EDA in those contexts. Future research should explore the use of EDA in a broader range of assembly operations to validate its generalizability and effectiveness in various industrial applications.

\appendices

\section{Categories and parameter ranges}\label{ap:Categories}
Table \ref{tab:Categories} shows three different categories for each of the parameter ranges. The units for the translational stiffnesses and rotational stiffnesses are N/m and Nm/rad, respectively. The damping ratio is unitless.
\label{tab:Categories}
\begin{tabular}{@{}cccc@{}}
\toprule
Category      & $k_t$ & $k_r$ & $\zeta$ \\ \midrule
Small & $50 \leq k_t \leq 300$ & $5 \leq k_r \leq 10$ & $0.1 \leq \zeta \leq 0.3$ \\
Medium   & $301 \leq k_t \leq 700$ & $11 \leq k_r \leq 80$ & $0.4 \leq \zeta \leq 0.7$ \\
Large  & $701 \leq k_t \leq 1000$ & $81 \leq k_r \leq 200$ & $0.8 \leq \zeta \leq 0.9$ \\ \bottomrule
\end{tabular}

\section{Principal Component Analysis}\label{ap:PCA}

\begin{figure*}[htp]
\includegraphics[trim=4.5cm 0 0 0cm, width = 1.1\textwidth]{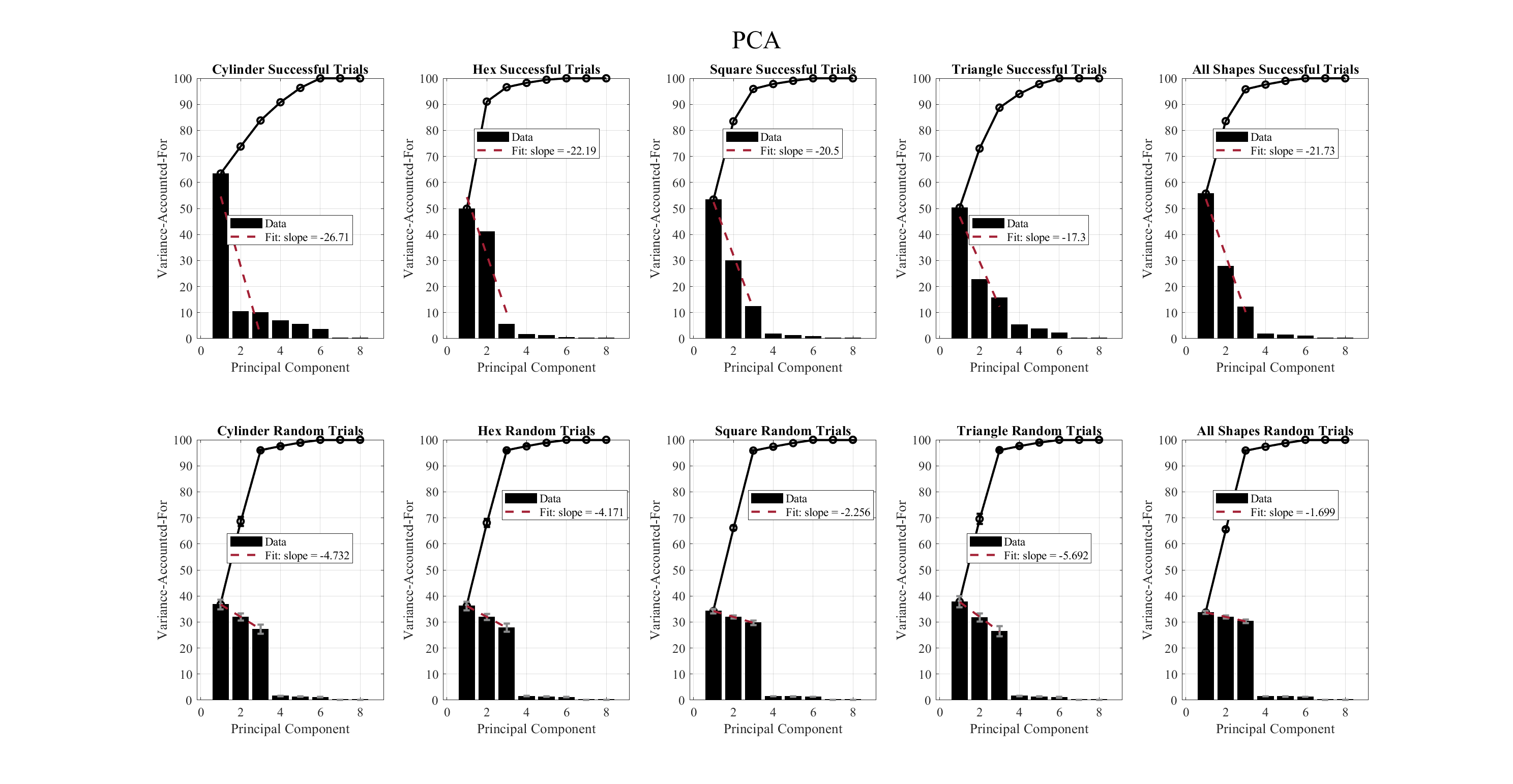}
\centering
\caption{\textbf{Top Row}: Variance-accounted-for of the successful trials after Principal Component Analysis. \textbf{Bottom Row}: Variance-accounted-for of a random subset of trials after Principal Component Analysis. 
The black line represents a cumulative variance-accounted-for. The dashed red line represents a linear fit to the first three principal components; its slope is written in the legend.}
\label{fig:VAF_slope}
\end{figure*}

We conducted a pair-wise t-test between slopes of the variance-accounted-for in the successful trials vs. the random trials. If the slopes in the successful trials were significantly steeper (i.e., the probability that the slopes were the same was  $\alpha \leq 0.05$), we concluded that the PCs extracted from the successful trials were not a  random artifact. Specifically, we concluded that there existed a \textit{non-arbitrary} impedance solution subspace. The results can be seen in Fig.~\ref{fig:VAF_slope}.

\section{Low Dimensional Representation of Solutions}\label{ap:projections}
\begin{figure*}[htp]
\includegraphics[width = \textwidth]{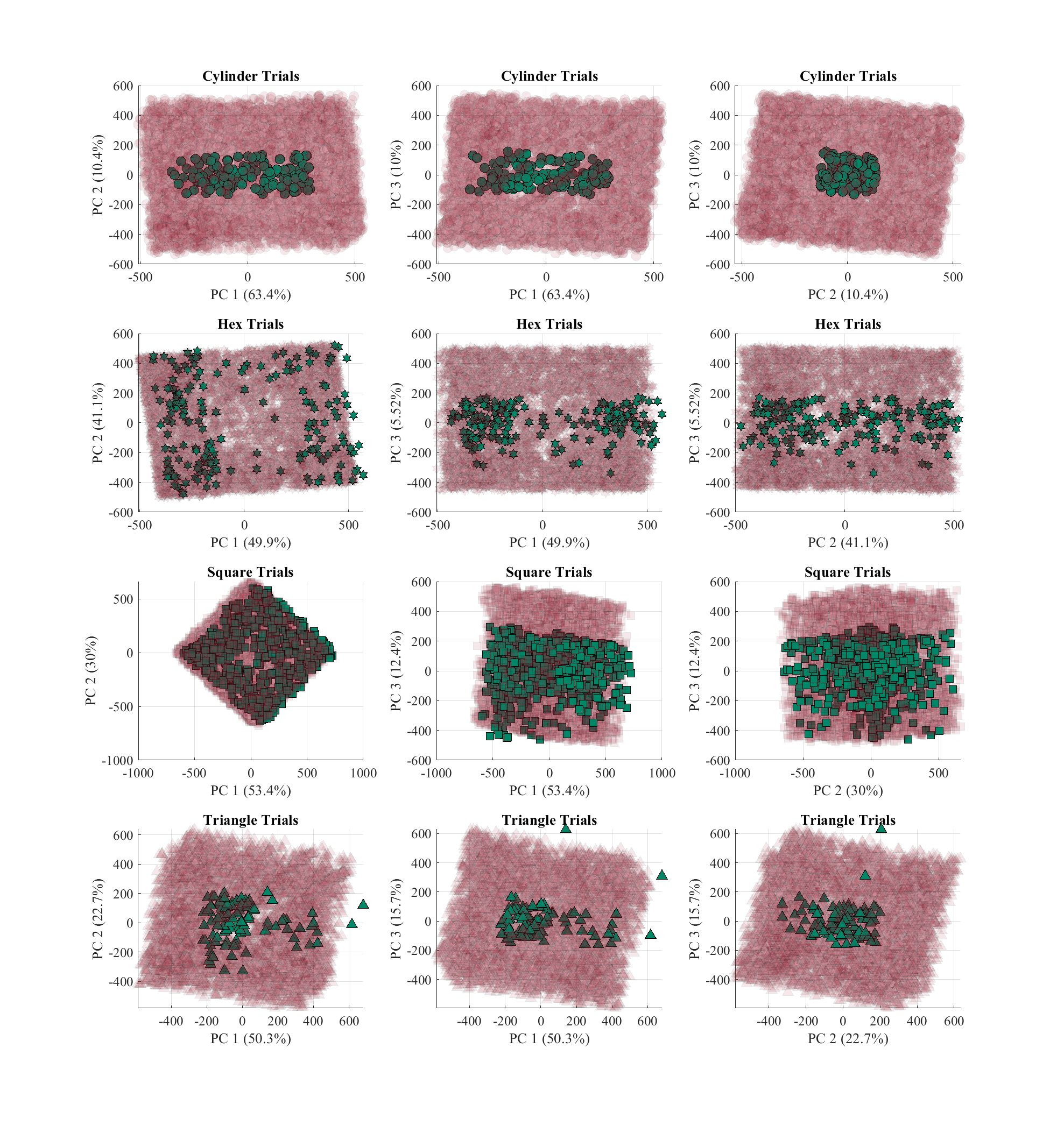}
\centering
\caption{\textbf{2D-projections} of the successful (green) and unsuccessful (red) trials for all pegs projected onto the solution space.}
\label{fig:SuccessvsFailure_combined_projections}
\end{figure*}

Figure~\ref{fig:SuccessvsFailure_combined_projections} shows 2D-projections of the successful (green) and unsuccessful (red) trials of the cylinder, hex, square, and triangle pegs are projected into the solution space obtained using PCA (from \ref{fig:SuccessvsFailure_combined}). Note that often the failures fell outside the PCA-reduced solution space.

\section{Optimal Cluster Count Using Silhouette Analysis}\label{ap:SilhouetteCoeff}

The silhouette value  $s_i$  measures how well a data point $i$ fits within its assigned cluster compared to other clusters. It is computed using the formula:
\begin{equation}
    s_i = \sum_{i=1}^N \frac{b_i - a_i}{max(a_i,b_i)},
\end{equation}
where $a_i$ is the average distance between the $i-$th point and all other data points within the cluster, and $b_i$ is the avg distance between the $i-$th point and all other data points within the next nearest cluster. From the above definition it is clear that $-1 \leq s_i \leq 1$. Values close to 1 indicate well-clustered data, negative values suggest misclustered data, and values near zero indicate the datum is near the border of two clusters \cite{rousseeuw1987silhouettes,kaufman2009finding}. 

The mean $s_i$ over all points of a cluster is a measure of how tightly grouped all the points in the cluster are. Thus, the mean $s_i$ over the entire dataset is a measure of how appropriately the data have been clustered. If there are too many or too few clusters, as may occur when a poor choice of $k$ is used in the clustering algorithm, some of the clusters will typically display much lower silhouette values than the rest. 

With this, the overall clustering effectiveness can be evaluated by the silhouette score, $SC$, defined as:
\begin{equation}
    SC = \max_{k} s(k),
\end{equation}
where $k$ represents the number of clusters and $s(k)$ represents the mean over all $s_i$ of the entire dataset for that number of clusters $k$. The number of clusters $k$ that maximizes the silhouette coefficient $SC$ is the most effective number of clusters for the k-means clustering algorithm \cite{rousseeuw1987silhouettes,kaufman2009finding}.

\subsection{Silhouette coefficients by peg type}\label{ap:SilhouetteCoeff_pegTypes}
\begin{figure}[h]
\includegraphics[width = \columnwidth]{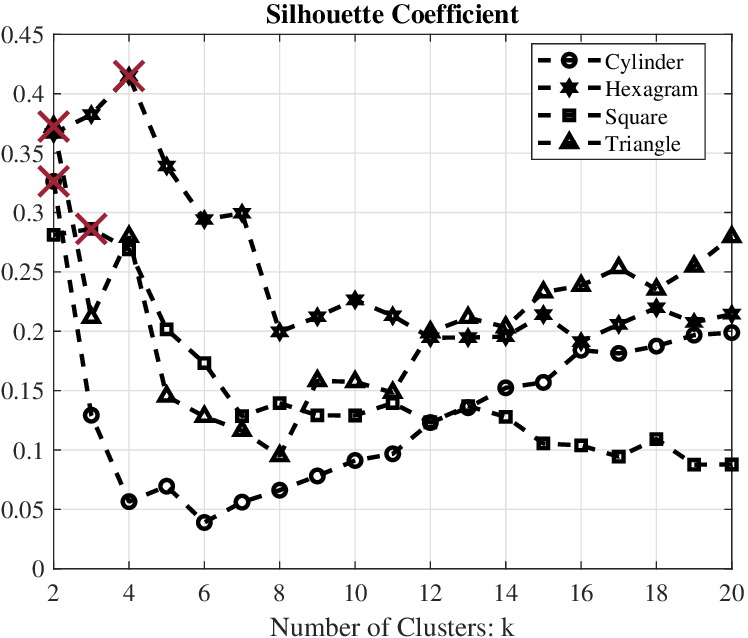}
\centering
\caption{The silhouette coefficient found from k-means clustering as a function of the number of clusters. The optimal number of clusters for each peg is denoted by the red X.}
\label{fig:SilhouetteCoefficent}
\end{figure}

When using k-means clustering to identify families of solutions in the cylinder, hex, square, and triangle pegs we found 2, 4, 3, and 2 solution clusters, respectively. These clusters had silhouette coefficients of 0.33, 0.41, 0.29, and 0.37 for the cylinder, hex, square, and triangle pegs, as can be seen in Fig.~\ref{fig:SilhouetteCoefficent}.

\section{MANOVA to identify distinct clusters}\label{ap:MANOVA}
To determine if the identified clusters were indeed distinct, we conducted a one-way Multivariate Analysis of Variance (MANOVA). MANOVA tests the null hypothesis that the means of each group are the same $n$-dimensional multivariate vector and that any observed differences in the sample are due to random chance. Note that the null hypothesis is examined across each of the $n$ dimensions where the largest possible dimension is either the dimension of the space or one less than the number of groups. Specifically, we used MANOVA to assess the effect of cluster group on the eight impedance parameters (dependent variables) for different shapes (cylinder, hex, square, and triangle). The independent variable was the cluster group, which had varying levels depending on the number of clusters identified for a given shape. The MANOVA was implemented using the \texttt{manova1} function in MATLAB \cite{krzanowski2000principles}. This function returns an estimate of the dimension of the space containing the group means and tests the null hypothesis that the group means lie within a subspace of the given dimension. Specifically, it assesses whether the means of the groups can be distinguished when projected into subspaces of increasing dimensionality, ranging from dimension zero (indicating no significant difference between the group means) up to dimension $k-1$, where $k$ is the number of groups. The upper limit of $k-1$ dimensions arises because there are $k-1$ degrees of freedom in comparing $k$ groups. The results from the \texttt{manova1} function were used to determine the significance of the differences among the clusters.

\subsection{MANOVA combined results}\label{ap:MANOVA_combined}

The MANOVA results revealed significant differences among the clusters on the combined dependent variables, suggesting that the clusters' multivariate means occupied at least three dimensions:

\begin{itemize}
    \item Dimension 0: Wilks' Lambda $= 0.0315$, $\chi^2(24) = 3693.09$, $p < 0.001$
    \item Dimension 1: Wilks' Lambda $= 0.2383$, $\chi^2(14) = 1,531.56$, $p < 0.001$
    \item Dimension 2: Wilks' Lambda $= 0.9824$, $\chi^2(6) = 18.91$, $p = 0.0043$
\end{itemize}

\subsection{MANOVA results by peg type}\label{ap:MANOVA_pegTypes}
Cylinder: The independent variable (cluster group) had two levels. MANOVA results showed that the multivariate means of the clusters occupied a one-dimensional space:
\begin{itemize}
    \item Wilks' Lambda $= 0.2198$, $\chi^2(8) = 213.62$, $p < 0.001$
\end{itemize}

\noindent Hex: The independent variable had four levels. The clusters spanned at least three dimensions:
\begin{itemize}
    \item Dimension 0: Wilks' Lambda $= 0.0197$, $\chi^2(24) = 683.51$, $p < 0.001$
    \item Dimension 1: Wilks' Lambda $= 0.1854$, $\chi^2(14) = 293.27$, $p < 0.001$
    \item Dimension 2: Wilks' Lambda $= 0.9087$, $\chi^2(6) = 16.65$, $p = 0.0106$
\end{itemize}

\noindent Square: The independent variable had three levels. The clusters spanned at least two dimensions:
\begin{itemize}
    \item Dimension 0: Wilks' Lambda $= 0.1029$, $\chi^2(16) = 1454.05.39$, $p < 0.001$
    \item Dimension 1: Wilks' Lambda $= 0.4322$, $\chi^2(7) = 536.42$, $p < 0.001$
\end{itemize}

\noindent Triangle: The independent variable had two levels. MANOVA results showed a one-dimensional space:
\begin{itemize}
    \item Wilks' Lambda $= 0.2391$, $\chi^2(8) = 135.91$, $p < 0.001$
\end{itemize}

\section{Neural Network structure}\label{ap:NeuralNetwork}
For the neural network, we chose the following structure with two hidden layers:
\begin{equation}
    f(\bm{x}) = f_3(\bm{W}_3 \ f_2(\bm{W}_2 \ f_1(\bm{W}_1 \bm{x} + \bm{b}_1) + \bm{b}_2) + b_3).
\end{equation}
Here, $\bm{W}_1, \bm{W}_2, \bm{W}_3$ are the weights of the network, $f_1, f_2, f_3$ are activation functions, and $b_1, b_2, b_3$ are bias terms associated with each neuron in the network.


The two hidden layers and output of the neural network were defined as follows: the first hidden layer transformed the eight-dimensional normalized input features $\mathbf{X}^{(i)}=\{k_x^{(i)}, k_y^{(i)}, k_z^{(i)}, k_C^{(i)}, k_B^{(i)}, k_A^{(i)}, \zeta_t^{(i)}, \zeta_r^{(i)} \}\in\mathbb{R}_{>0}^{8}$ to 64 neurons; the second hidden layer took the 64 features from the hidden layer and passed them to another 64 neurons; the output layer transformed the 64 outputs of the second layer to the predicted output $\hat{y}^{(i)}$. After each of the first two layers, a Rectified Linear Unit (ReLU) activation function was used to introduce non-linearity into the model. A Sigmoid activation function was used at the output layer to map the final value to a probability between 0 and 1. The Adam optimizer \cite{kingma2014adam} was used to update the weights of the network with a fixed learning rate of $1\times10^{-4}$.

\section{Linear Model Fitting Performance for Impedance Parameters}\label{ap:LinearRegression}
Tab.~\ref{tab:succ_rate_test} presents the complete details of the linear model results. Moreover, Tab.~\ref{tab:lin_fit_stats} reports the coefficients of determination and related p-values for the linear models used to fit the relative success rates of Fig.~\ref{fig:relative_succ_rate}.

\begin{table*}[htp]
\centering
\caption{Linear model fitting performance of the success rate distribution for four impedance parameters across all  pegs.  $R^2$ is the coefficient of determination of the linear fit. The p-value is the outcome of a statistical test of the linear model against the null slope. Asterisks denote significant differences.}
\label{tab:succ_rate_test}
\begin{tabular}{ccccccccc}
\hline
\multirow{2}{*}{Peg} & \multicolumn{2}{c}{$k_z$}   & \multicolumn{2}{c}{$k_C$}   & \multicolumn{2}{c}{$\zeta_t$} & \multicolumn{2}{c}{$\zeta_r$} \\ \cline{2-9} 
                     & $R^2$   & p-value           & $R^2$   & p-value           & $R^2$          & p-value       & $R^2$          & p-value       \\ \hline
Triangle             & 0.86 & \textless{}0.001* & 0.73 & 0.002*            & 0.10        & 0.501         & 0.52        & 0.066         \\
Square               & 0.91 & \textless{}0.001* & 0.85 & \textless{}0.001* & 0.30        & 0.201         & 0.73        & 0.015*        \\
Hexagon              & 0.77 & \textless{}0.001* & 0.94 & \textless{}0.001* & 0.58        & 0.048*        & 0.26        & 0.247         \\
Cylinder             & 0.65 & 0.005*            & 0.88 & \textless{}0.001* & 0.14        & 0.403         & 0.09        & 0.524         \\ \hline
\end{tabular}
\end{table*}

\begin{table*}[htp]
\centering
\caption{Coefficients of determination ($R^2$) and p-values of the linear fit of each relative success histogram for the four different pegs. Asterisks (*) denote significant difference between the linear models and the degenerate null slope.}
\label{tab:lin_fit_stats}
\begin{tabular}{l|ll|ll|ll|ll}
      & \multicolumn{2}{l}{Triangle} & \multicolumn{2}{l}{Square} & \multicolumn{2}{l}{Hexagon} & \multicolumn{2}{l}{Cylinder} \\
      & $R^2$           & p-value             & $R^2$          & p-value            & $R^2$          & p-value             & $R^2$           & p-value             \\ \hline
$k_x$    & 0.64         & 0.005*        & 0.60        & 0.009*       & 0.08        & 0.432         & 0.64         & 0.006*        \\
$k_y$    & 0.77         & $<0.001* $       & 0.32        & 0.088*       & 0.02        & 0.663         & 0.65         & 0.005*        \\
$k_z$    & 0.86         & $<0.001*$        & 0.91        & $<0.001*$       & 0.77        & $<0.001* $       & 0.65         & 0.005*        \\
$k_A$(z) & 0.54         & 0.016*        & 0.22        & 0.174        & 0.01        & 0.804         & 0.52         & 0.019*        \\
$k_B$(y) & 0.42         & 0.043*        & 0.48        & 0.026*       & 0.14        & 0.281         & 0.05         & 0.542         \\
$k_C$(x) & 0.73         & 0.002*        & 0.85        & $<0.001* $      & 0.94        & $<0.001*$        & 0.88         & $<0.001* $       \\
$\zeta_t$    & 0.10         & 0.501         & 0.30        & 0.201        & 0.58        & 0.048*        & 0.14         & 0.403         \\
$\zeta_r$    & 0.52         & 0.066         & 0.73        & 0.015*       & 0.26        & 0.247         & 0.09         & 0.524        
\end{tabular}
\end{table*}

\subsection{Success rates for individual neural networks}\label{ap:LinReg_Neural}
\begin{figure*}[htp]
\includegraphics[width = \textwidth]{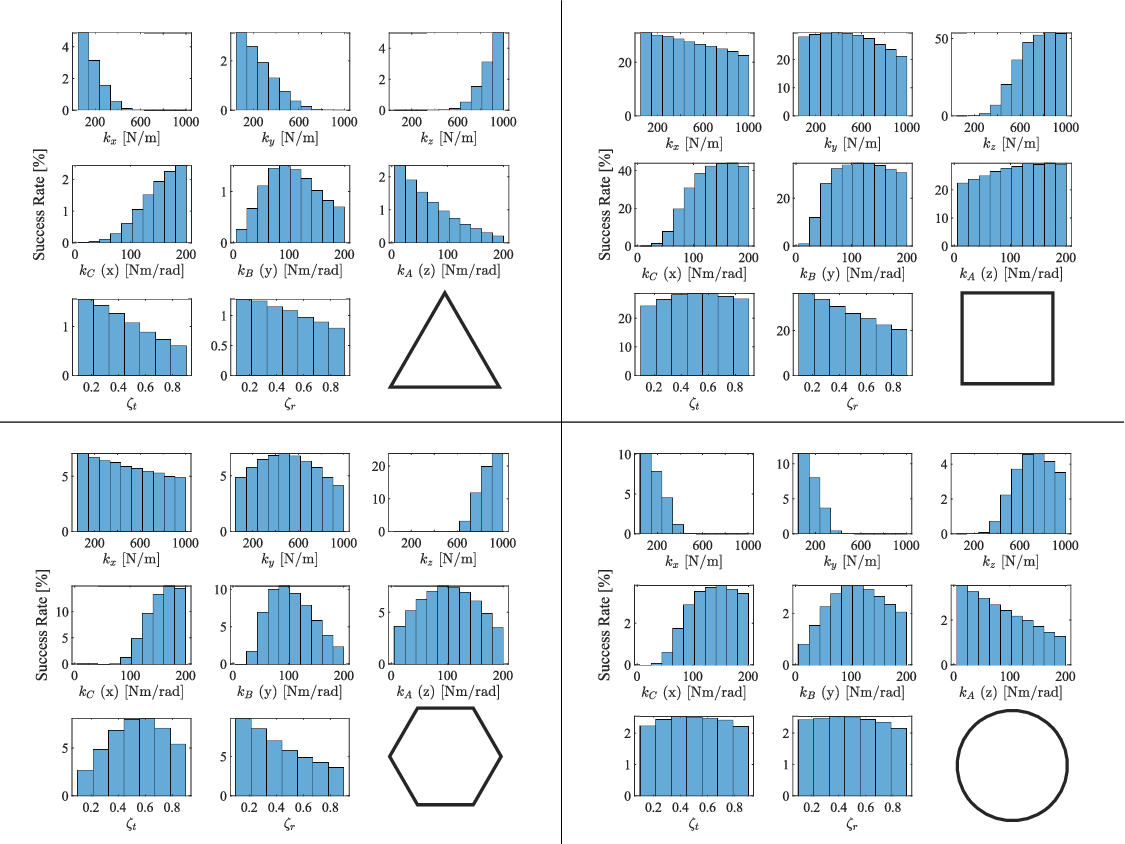}
\centering
\caption{Distribution of success rates for the parameters that were predicted to be successful according to the trained neural network. The distributions were generated by uniformly sampling impedance parameters between the prescribed minimum and maximum values presented in Sec.~\ref{sec:Methods} until 100,000 successes were generated. The resulting success rate (number of successful parameters over number of tested parameters) for each peg were: triangle $s_r = 1.02\%$, square $s_r = 27.07\%$, hexagon $s_r = 5.87 \%$, cylinder $s_r = 2.36\%$.}
\label{fig:nn_all_pegs}
\end{figure*}

Fig. \ref{fig:nn_all_pegs} shows the distribution of success rates for 4 different neural networks specifically trained on each individual peg: triangle, square, hexagon, and cylinder. The individual neural networks shared the same structure as the neural network presented in the main manuscript. They were trained using the same split (70/30) between training and validation data. The resulting accuracy of the neural networks were respectively: triangle $98.53\%$, square $95.84\%$, hexagon $97.41\%$, and cylinder $98.83\%$. 
The success rate distributions were generated by uniformly sampling the 8 impedance parameters between their maximum and minimum values until 100,000 successes were collected. The resulting success rate for each peg was: square $s_r = 27.07\%$, hexagon $s_r = 5.87 \%$, triangle $s_r = 1.02\%$, cylinder $s_r = 2.36\%$.
It is interesting to observe how in some cases (square, hexagon) the success rate of the neural network was higher than the experimental data, while in the other cases (triangle, cylinder) the success rate was practically indistinguishable from the experimental data.

\section*{Acknowledgment}
This work was supported in part by the MIT/SUSTech Centers for Mechanical Engineering Research and Education. 
MCN was supported in part by a Mathworks Fellowship. 
JL was supported in part by the MIT-Novo Nordisk Artificial Intelligence Postdoctoral Fellows Program. AMWJ was supported by the Accenture Fellowship.

We gratefully acknowledge the support of KUKA, an international leader in automation solutions, and specifically thank them for providing the KUKA robots used in the robotic experiments.

\bibliographystyle{IEEEtran}
\bibliography{literature}

\begin{thebibliography}{10}
\providecommand{\url}[1]{#1}
\csname url@samestyle\endcsname
\providecommand{\newblock}{\relax}
\providecommand{\bibinfo}[2]{#2}
\providecommand{\BIBentrySTDinterwordspacing}{\spaceskip=0pt\relax}
\providecommand{\BIBentryALTinterwordstretchfactor}{4}
\providecommand{\BIBentryALTinterwordspacing}{\spaceskip=\fontdimen2\font plus
\BIBentryALTinterwordstretchfactor\fontdimen3\font minus
  \fontdimen4\font\relax}
\providecommand{\BIBforeignlanguage}[2]{{%
\expandafter\ifx\csname l@#1\endcsname\relax
\typeout{** WARNING: IEEEtran.bst: No hyphenation pattern has been}%
\typeout{** loaded for the language `#1'. Using the pattern for}%
\typeout{** the default language instead.}%
\else
\language=\csname l@#1\endcsname
\fi
#2}}
\providecommand{\BIBdecl}{\relax}
\BIBdecl

\bibitem{nagabandi2020deep}
A.~Nagabandi, K.~Konolige, S.~Levine, and V.~Kumar, ``Deep dynamics models for
  learning dexterous manipulation,'' in \emph{Conference on Robot
  Learning}.\hskip 1em plus 0.5em minus 0.4em\relax PMLR, 2020, pp. 1101--1112.

\bibitem{andrychowicz2020learning}
O.~M. Andrychowicz, B.~Baker, M.~Chociej, R.~Jozefowicz, B.~McGrew,
  J.~Pachocki, A.~Petron, M.~Plappert, G.~Powell, A.~Ray \emph{et~al.},
  ``Learning dexterous in-hand manipulation,'' \emph{The International Journal
  of Robotics Research}, vol.~39, no.~1, pp. 3--20, 2020.

\bibitem{chen2022system}
T.~Chen, J.~Xu, and P.~Agrawal, ``A system for general in-hand object
  re-orientation,'' in \emph{Conference on Robot Learning}.\hskip 1em plus
  0.5em minus 0.4em\relax PMLR, 2022, pp. 297--307.

\bibitem{seo2023contact}
J.~Seo, N.~P. Prakash, X.~Zhang, C.~Wang, J.~Choi, M.~Tomizuka, and
  R.~Horowitz, ``Contact-rich se (3)-equivariant robot manipulation task
  learning via geometric impedance control,'' \emph{IEEE Robotics and
  Automation Letters}, 2023.

\bibitem{pang2023global}
T.~Pang, H.~T. Suh, L.~Yang, and R.~Tedrake, ``Global planning for contact-rich
  manipulation via local smoothing of quasi-dynamic contact models,''
  \emph{IEEE Transactions on Robotics}, 2023.

\bibitem{xie2023neural}
M.~Xie, A.~Handa, S.~Tyree, D.~Fox, H.~Ravichandar, N.~D. Ratliff, and
  K.~Van~Wyk, ``Neural geometric fabrics: Efficiently learning high-dimensional
  policies from demonstration,'' in \emph{Conference on Robot Learning}.\hskip
  1em plus 0.5em minus 0.4em\relax PMLR, 2023, pp. 1355--1367.

\bibitem{posa2014direct}
M.~Posa, C.~Cantu, and R.~Tedrake, ``A direct method for trajectory
  optimization of rigid bodies through contact,'' \emph{The International
  Journal of Robotics Research}, vol.~33, no.~1, pp. 69--81, 2014.

\bibitem{manchester2020variational}
Z.~Manchester and S.~Kuindersma, ``Variational contact-implicit trajectory
  optimization,'' in \emph{Robotics Research: The 18th International Symposium
  ISRR}.\hskip 1em plus 0.5em minus 0.4em\relax Springer, 2020, pp. 985--1000.

\bibitem{oikawa2021reinforcement}
M.~Oikawa, T.~Kusakabe, K.~Kutsuzawa, S.~Sakaino, and T.~Tsuji, ``Reinforcement
  learning for robotic assembly using equivalent-diagonal stiffness matrix,''
  \emph{IEEE Robotics and Automation Letters}, vol.~6, no.~2, pp. 2737--2744,
  2021.

\bibitem{Chi_2023_Diffusion}
C.~Chi, S.~Feng, Y.~Du, Z.~Xu, E.~Cousineau, B.~Burchfiel, and S.~Song,
  ``Diffusion policy: Visuomotor policy learning via action diffusion,''
  \emph{arXiv preprint arXiv:2303.04137}, 2023.

\bibitem{marcucci2017approximate}
T.~Marcucci, R.~Deits, M.~Gabiccini, A.~Bicchi, and R.~Tedrake, ``Approximate
  hybrid model predictive control for multi-contact push recovery in complex
  environments,'' in \emph{2017 IEEE-RAS 17th international conference on
  humanoid robotics (Humanoids)}.\hskip 1em plus 0.5em minus 0.4em\relax IEEE,
  2017, pp. 31--38.

\bibitem{hogan2020reactive}
F.~R. Hogan and A.~Rodriguez, ``Reactive planar non-prehensile manipulation
  with hybrid model predictive control,'' \emph{The International Journal of
  Robotics Research}, vol.~39, no.~7, pp. 755--773, 2020.

\bibitem{anitescu1997formulating}
M.~Anitescu and F.~A. Potra, ``Formulating dynamic multi-rigid-body contact
  problems with friction as solvable linear complementarity problems,''
  \emph{equivalentlinear Dynamics}, vol.~14, pp. 231--247, 1997.

\bibitem{aydinoglu2022real}
A.~Aydinoglu and M.~Posa, ``Real-time multi-contact model predictive control
  via admm,'' in \emph{2022 International Conference on Robotics and Automation
  (ICRA)}.\hskip 1em plus 0.5em minus 0.4em\relax IEEE, 2022, pp. 3414--3421.

\bibitem{theodorou2010generalized}
E.~Theodorou, J.~Buchli, and S.~Schaal, ``A generalized path integral control
  approach to reinforcement learning,'' \emph{The Journal of Machine Learning
  Research}, vol.~11, pp. 3137--3181, 2010.

\bibitem{kalakrishnan2011stomp}
M.~Kalakrishnan, S.~Chitta, E.~Theodorou, P.~Pastor, and S.~Schaal, ``Stomp:
  Stochastic trajectory optimization for motion planning,'' in \emph{2011 IEEE
  international conference on robotics and automation}.\hskip 1em plus 0.5em
  minus 0.4em\relax IEEE, 2011, pp. 4569--4574.

\bibitem{buchli2011learning}
J.~Buchli, F.~Stulp, E.~Theodorou, and S.~Schaal, ``Learning variable impedance
  control,'' \emph{The International Journal of Robotics Research}, vol.~30,
  no.~7, pp. 820--833, 2011.

\bibitem{stulp2012reinforcement}
F.~Stulp, E.~A. Theodorou, and S.~Schaal, ``Reinforcement learning with
  sequences of motion primitives for robust manipulation,'' \emph{IEEE
  Transactions on robotics}, vol.~28, no.~6, pp. 1360--1370, 2012.

\bibitem{williams2017model}
G.~Williams, A.~Aldrich, and E.~A. Theodorou, ``Model predictive path integral
  control: From theory to parallel computation,'' \emph{Journal of Guidance,
  Control, and Dynamics}, vol.~40, no.~2, pp. 344--357, 2017.

\bibitem{abraham2020model}
I.~Abraham, A.~Handa, N.~Ratliff, K.~Lowrey, T.~D. Murphey, and D.~Fox,
  ``Model-based generalization under parameter uncertainty using path integral
  control,'' \emph{IEEE Robotics and Automation Letters}, vol.~5, no.~2, pp.
  2864--2871, 2020.

\bibitem{van2023learning}
M.~Van~der Merwe, D.~Berenson, and N.~Fazeli, ``Learning the dynamics of
  compliant tool-environment interaction for visuo-tactile contact servoing,''
  in \emph{Conference on Robot Learning}.\hskip 1em plus 0.5em minus
  0.4em\relax PMLR, 2023, pp. 2052--2061.

\bibitem{suh2022bundled}
H.~J.~T. Suh, T.~Pang, and R.~Tedrake, ``Bundled gradients through contact via
  randomized smoothing,'' \emph{IEEE Robotics and Automation Letters}, vol.~7,
  no.~2, pp. 4000--4007, 2022.

\bibitem{suh2022differentiable}
H.~J. Suh, M.~Simchowitz, K.~Zhang, and R.~Tedrake, ``Do differentiable
  simulators give better policy gradients?'' in \emph{International Conference
  on Machine Learning}.\hskip 1em plus 0.5em minus 0.4em\relax PMLR, 2022, pp.
  20\,668--20\,696.

\bibitem{duchi2012randomized}
J.~C. Duchi, P.~L. Bartlett, and M.~J. Wainwright, ``Randomized smoothing for
  stochastic optimization,'' \emph{SIAM Journal on Optimization}, vol.~22,
  no.~2, pp. 674--701, 2012.

\bibitem{curtis2012sequential}
F.~E. Curtis and M.~L. Overton, ``A sequential quadratic programming algorithm
  for nonconvex, nonsmooth constrained optimization,'' \emph{SIAM Journal on
  Optimization}, vol.~22, no.~2, pp. 474--500, 2012.

\bibitem{burke2020gradient}
J.~V. Burke, F.~E. Curtis, A.~S. Lewis, M.~L. Overton, and L.~E. Sim{\~o}es,
  ``Gradient sampling methods for nonsmooth optimization,'' \emph{Numerical
  nonsmooth optimization: State of the art algorithms}, pp. 201--225, 2020.

\bibitem{Manschitz_2020}
S.~Manschitz, M.~Gienger, J.~Kober, and J.~Peters, ``Learning sequential force
  interaction skills,'' \emph{Robotics}, vol.~9, no.~2, 2020.

\bibitem{schaal1999imitation}
S.~Schaal, ``Is imitation learning the route to humanoid robots?'' \emph{Trends
  in cognitive sciences}, vol.~3, no.~6, pp. 233--242, 1999.

\bibitem{Ijspeert_2013}
A.~J. Ijspeert, J.~Nakanishi, H.~Hoffmann, P.~Pastor, and S.~Schaal,
  ``Dynamical movement primitives: learning attractor models for motor
  behaviors,'' \emph{Neural computation}, vol.~25, no.~2, pp. 328--373, 2013.

\bibitem{hogan2012dynamic}
N.~Hogan and D.~Sternad, ``Dynamic primitives of motor behavior,''
  \emph{Biological cybernetics}, vol. 106, no.~11, pp. 727--739, 2012.

\bibitem{nah2024MotorPrimitives}
C.~Nah, J.~Lachner, and N.~Hogan, ``\BIBforeignlanguage{en}{Robot control based
  on motor primitives - a comparison of two approaches},''
  \emph{\BIBforeignlanguage{en}{The International Journal of Robotics
  Research}}, 2024.

\bibitem{kim2011impedance}
B.-S. Kim, Y.-L. Kim, J.-B. Song, and S.-W. Son, ``Impedance-control based
  peg-in-hole assembly with a 6 dof manipulator,'' \emph{Transactions of the
  Korean Society of Mechanical Engineers A}, vol.~35, no.~4, pp. 347--352,
  2011.

\bibitem{park2013intuitive}
H.~Park, J.-H. Bae, J.-H. Park, M.-H. Baeg, and J.~Park, ``Intuitive
  peg-in-hole assembly strategy with a compliant manipulator,'' in \emph{IEEE
  ISR 2013}.\hskip 1em plus 0.5em minus 0.4em\relax IEEE, 2013, pp. 1--5.

\bibitem{sherrington1906integrative}
C.~S. Sherrington, \emph{The integrative action of the nervous system}.\hskip
  1em plus 0.5em minus 0.4em\relax A. Constable, 1906.

\bibitem{bernstein_1967}
\BIBentryALTinterwordspacing
N.~Bernstein, \emph{The Co-ordination and Regulation of Movements}.\hskip 1em
  plus 0.5em minus 0.4em\relax Pergamon Press, 1967. [Online]. Available:
  \url{https://books.google.com/books?id=mUhzjwEACAAJ}
\BIBentrySTDinterwordspacing

\bibitem{Latash_2012}
M.~Latash, ``The bliss (not the problem) of motor abundance (not redundancy),''
  \emph{Experimental brain research. Experimentelle Hirnforschung.
  Expérimentation cérébrale}, vol. 217, pp. 1--5, 03 2012.

\bibitem{wolpert1998multiple}
D.~M. Wolpert and M.~Kawato, ``Multiple paired forward and inverse models for
  motor control,'' \emph{Neural networks}, vol.~11, no. 7-8, pp. 1317--1329,
  1998.

\bibitem{Wolpert_2001}
\BIBentryALTinterwordspacing
D.~M. Wolpert, Z.~Ghahramani, and J.~R. Flanagan, ``Perspectives and problems
  in motor learning,'' \emph{Trends in Cognitive Sciences}, vol.~5, pp.
  487--494, 2001. [Online]. Available:
  \url{https://api.semanticscholar.org/CorpusID:6351794}
\BIBentrySTDinterwordspacing

\bibitem{Lachner_2023_EDA}
\BIBentryALTinterwordspacing
J.~Lachner, M.~C. Nah, F.~Tessari, and N.~Hogan, ``Elementary dynamic actions:
  key structures for contact-rich manipulation,'' in \emph{IROS 2023 Workshop
  on Leveraging Models for Contact-Rich Manipulation}, 2023. [Online].
  Available: \url{https://openreview.net/forum?id=z6uHTf02KK}
\BIBentrySTDinterwordspacing

\bibitem{Nah_2023_EDA_kin}
M.~C. Nah, J.~Lachner, F.~Tessari, and N.~Hogan, ``Kinematic modularity of
  elementary dynamic actions,'' in \emph{2024 IEEE/RSJ International Conference
  on Intelligent Robots and Systems (IROS)}.\hskip 1em plus 0.5em minus
  0.4em\relax IEEE, 2024.

\bibitem{fasse1996control}
E.~D. Fasse and N.~Hogan, ``Control of physical contact and dynamic
  interaction,'' in \emph{Robotics Research: The Seventh International
  Symposium}.\hskip 1em plus 0.5em minus 0.4em\relax Springer, 1996, pp.
  28--38.

\bibitem{hogan1984impedance}
N.~Hogan, ``Impedance control: An approach to manipulation,'' in \emph{1984
  American control conference}.\hskip 1em plus 0.5em minus 0.4em\relax IEEE,
  1984, pp. 304--313.

\bibitem{hogan1985impedance}
------, ``Impedance control-an approach to manipulation. i-theory.
  ii-implementation. iii-applications,'' \emph{ASME Journal of Dynamic Systems
  and Measurement Control B}, vol. 107, pp. 1--24, 1985.

\bibitem{hogan2014general}
------, ``A general actuator model based on nonlinear equivalent networks,''
  \emph{IEEE/ASME Transactions on Mechatronics}, vol.~19, no.~6, pp.
  1929--1939, 2014.

\bibitem{hogan2018impedance}
N.~Hogan and S.~P. Buerger, ``Impedance and interaction control,'' in
  \emph{Robotics and automation handbook}.\hskip 1em plus 0.5em minus
  0.4em\relax CRC press, 2018, pp. 375--398.

\bibitem{hogan2017physical}
N.~Hogan, ``Physical interaction via dynamic primitives,'' in \emph{Geometric
  and numerical foundations of movements}.\hskip 1em plus 0.5em minus
  0.4em\relax Springer, 2017, pp. 269--299.

\bibitem{lachner2022geometric}
J.~Lachner, ``A geometric approach to robotic manipulation in physical
  human-robot interaction,'' Ph.D. dissertation, University of Twente, 2022.

\bibitem{hogan2013dynamic}
N.~Hogan and D.~Sternad, ``Dynamic primitives in the control of locomotion,''
  \emph{Frontiers in computational neuroscience}, vol.~7, p.~71, 2013.

\bibitem{hogan2022contact}
N.~Hogan, ``Contact and physical interaction,'' \emph{Annual Review of Control,
  Robotics, and Autonomous Systems}, vol.~5, pp. 179--203, 2022.

\bibitem{lachner_energy_2021}
J.~Lachner, F.~Allmendinger, E.~Hobert, N.~Hogan, and S.~Stramigioli,
  ``\BIBforeignlanguage{en}{Energy budgets for coordinate invariant robot
  control in physical human–robot interaction},''
  \emph{\BIBforeignlanguage{en}{The International Journal of Robotics
  Research}}, vol.~40, no. 8-9, pp. 968--985, Aug. 2021.

\bibitem{lachner_shaping_2022}
J.~Lachner, F.~Allmendinger, S.~Stramigioli, and N.~Hogan, ``Shaping impedances
  to comply with constrained task dynamics,'' \emph{IEEE Transactions on
  Robotics}, vol.~38, no.~5, pp. 2750--2767, 2022.

\bibitem{hogan2007rhythmic}
N.~Hogan and D.~Sternad, ``On rhythmic and discrete movements: reflections,
  definitions and implications for motor control,'' \emph{Experimental brain
  research}, vol. 181, no.~1, pp. 13--30, 2007.

\bibitem{Hermus2023}
J.~Hermus, ``A dynamic primitives hypothesis: A descriptive model of human
  physical interaction,'' Ph.D. dissertation, Massachusetts Institute of
  Technology, 2023.

\bibitem{nah2020dynamic}
M.~C. Nah, A.~Krotov, M.~Russo, D.~Sternad, and N.~Hogan, ``Dynamic primitives
  facilitate manipulating a whip,'' in \emph{2020 8th IEEE RAS/EMBS
  International Conference for Biomedical Robotics and Biomechatronics
  (BioRob)}.\hskip 1em plus 0.5em minus 0.4em\relax IEEE, 2020, pp. 685--691.

\bibitem{nah2023learning}
------, ``Learning to manipulate a whip with simple primitive actions-a
  simulation study,'' \emph{iScience}, 2023.

\bibitem{lachner2023elementary}
\BIBentryALTinterwordspacing
J.~Lachner, M.~C. Nah, F.~Tessari, and N.~Hogan, ``Elementary dynamic actions:
  key structures for contact-rich manipulation,'' in \emph{IROS 2023 Workshop
  on Leveraging Models for Contact-Rich Manipulation}, 2023. [Online].
  Available: \url{https://openreview.net/forum?id=z6uHTf02KK}
\BIBentrySTDinterwordspacing

\bibitem{Bellman_1957_DynamicProgramming}
R.~Bellman, \emph{{Dynamic Programming}}.\hskip 1em plus 0.5em minus
  0.4em\relax Dover Publications, 1957.

\bibitem{west2023}
A.~M. West, F.~Tessari, and N.~Hogan, ``The study of complex manipulation via
  kinematic hand synergies: The effects of data pre-processing,'' in \emph{2023
  International Conference on Rehabilitation Robotics (ICORR)}, 2023, pp. 1--6.

\bibitem{lloyd1982least}
S.~Lloyd, ``Least squares quantization in pcm,'' \emph{IEEE transactions on
  information theory}, vol.~28, no.~2, pp. 129--137, 1982.

\bibitem{arthur2006k}
D.~Arthur and S.~Vassilvitskii, ``k-means++: The advantages of careful
  seeding,'' Stanford, Tech. Rep., 2006.

\bibitem{abu2020variable}
F.~J. Abu-Dakka and M.~Saveriano, ``Variable impedance control and learning—a
  review,'' \emph{Frontiers in Robotics and AI}, vol.~7, p. 590681, 2020.

\bibitem{martin2019variable}
R.~Mart{\'\i}n-Mart{\'\i}n, M.~A. Lee, R.~Gardner, S.~Savarese, J.~Bohg, and
  A.~Garg, ``Variable impedance control in end-effector space: An action space
  for reinforcement learning in contact-rich tasks,'' in \emph{2019 IEEE/RSJ
  International Conference on Intelligent Robots and Systems (IROS)}.\hskip 1em
  plus 0.5em minus 0.4em\relax IEEE, 2019, pp. 1010--1017.

\bibitem{Whitney_1982}
\BIBentryALTinterwordspacing
D.~E. Whitney, ``{Quasi-Static Assembly of Compliantly Supported Rigid
  Parts},'' \emph{Journal of Dynamic Systems, Measurement, and Control}, vol.
  104, no.~1, pp. 65--77, 03 1982. [Online]. Available:
  \url{https://doi.org/10.1115/1.3149634}
\BIBentrySTDinterwordspacing

\bibitem{Hirai1990remote}
S.~Hirai and K.~Tanie, ``Remote center compliance with adaptive compliance
  control,'' in \emph{Proceedings of the IEEE International Conference on
  Robotics and Automation}.\hskip 1em plus 0.5em minus 0.4em\relax IEEE, 1990,
  pp. 468--473.

\bibitem{rousseeuw1987silhouettes}
P.~J. Rousseeuw, ``Silhouettes: a graphical aid to the interpretation and
  validation of cluster analysis,'' \emph{Journal of computational and applied
  mathematics}, vol.~20, pp. 53--65, 1987.

\bibitem{kaufman2009finding}
L.~Kaufman and P.~J. Rousseeuw, \emph{Finding groups in data: an introduction
  to cluster analysis}.\hskip 1em plus 0.5em minus 0.4em\relax John Wiley \&
  Sons, 2009.

\bibitem{krzanowski2000principles}
W.~Krzanowski, \emph{Principles of multivariate analysis}.\hskip 1em plus 0.5em
  minus 0.4em\relax OUP Oxford, 2000, vol.~23.

\bibitem{kingma2014adam}
D.~P. Kingma and J.~Ba, ``Adam: A method for stochastic optimization,''
  \emph{arXiv preprint arXiv:1412.6980}, 2014.

\end{thebibliography}

\end{document}